\documentclass[1p]{elsarticle} 

\usepackage{hyperref}

\journal{Journal of Artificial Intelligence}

\usepackage{amsmath}
\usepackage{amsfonts}
\usepackage{xcolor}
\usepackage{verbatim}
\usepackage{paralist}
\usepackage{booktabs}
\usepackage{ifthen}
\usepackage{siunitx}
\usepackage{cleveref}
\usepackage{subcaption}
\usepackage{multirow}
\usepackage{cleveref}

\usepackage{my_macros}

\newcommand{\ShowComments}{true} 
\newcommand{\SR}[1]{
    \ifthenelse{\equal{\ShowComments}{true}}{\noindent\textcolor{red}{Sadegh: #1}}{}}
\newcommand{\JB}[1]{
    \ifthenelse{\equal{\ShowComments}{true}}{\noindent\textcolor{green}{Joydeep: #1}}{}}
\newcommand{\notes}[1]{
    \ifthenelse{\equal{\ShowComments}{true}}{\noindent\textcolor{blue}{#1}}{}}

\newcommand{\ShowTODO}{false} 
\newcommand{\TODO}[1]{
        \ifthenelse{\equal{\ShowTODO}{true}}{\noindent\textcolor{orange}{TODO: #1}}{}}

\newcommand{\HighlightRevisions}{false} 
\newcommand{\changes}[1]{
    \ifthenelse{\equal{\HighlightRevisions}{true}}{\textcolor{blue}{#1}}{\textcolor{black}{#1}}
}
\newcommand{\changesCaption}[1]{
    \ifthenelse{\equal{\HighlightRevisions}{true}}{\caption{\textcolor{blue}{#1}}}{\noindent\caption{\textcolor{black}{#1}}}
}

\begin{document}

\begin{frontmatter}

\title{Introspective Perception for Mobile Robots}

\author{Sadegh Rabiee \corref{mycorrespondingauthor}}
\cortext[mycorrespondingauthor]{Corresponding author}
\ead{srabiee@cs.utexas.edu}
\author{Joydeep Biswas\corref{}}
\ead{joydeepb@cs.utexas.edu}
\address{The Department of Computer
Science, The University of Texas at Austin, Austin TX 78712}

\begin{abstract}

Perception algorithms that provide estimates of their uncertainty are crucial to the development of autonomous robots that can operate in challenging and uncontrolled environments. Such perception algorithms provide the means for having risk-aware robots that reason about the probability of successfully completing a task when planning.
There exist perception algorithms that come with models of their uncertainty; however, these models are often developed with assumptions, such as perfect data associations, that do not hold in the real world. Hence the resultant estimated uncertainty is a weak lower bound.
To tackle this problem we present introspective perception --- a novel approach for predicting accurate estimates of the uncertainty of perception algorithms deployed on mobile robots. 
By exploiting sensing redundancy and consistency constraints naturally present in the data collected by a mobile robot, introspective perception learns an empirical model of the error distribution of perception algorithms in the deployment environment and in an autonomously supervised manner.
In this paper, we present the general theory of introspective perception and demonstrate successful implementations for two different perception tasks. We provide empirical results on challenging real-robot data for introspective stereo depth estimation and introspective visual simultaneous localization and mapping and show that they learn to predict their uncertainty with high accuracy and leverage this information to significantly reduce state estimation errors for an autonomous mobile robot.

\end{abstract}

\begin{keyword}
Competence-Aware Perception\sep Computer Vision \sep Introspection \sep Mobile Robots
\MSC[2010] 68T40\sep  68T45 \sep 68T37
\end{keyword}

\end{frontmatter}


\section{Introduction}\label{sec:introduction}

As robots become increasingly available, they are deployed for tasks where autonomous navigation in unstructured environments is required.
Robots deployed in challenging and previously unseen environments are likely to encounter failures due to situations that were not previously considered by robot developers.
One of the significant sources of failure in autonomy is errors in perception, hence, having perception algorithms that are capable of learning to predict the probability of errors in their output is of great importance.
Being able to predict the uncertainty in the output of perception empowers a robot to 1) improve state estimation when fusing information from several imperfect perception algorithms, 2) generate plans that reduce the probability of task failures, and 3) request human supervision efficiently and only when it is actually required.
Therefore, it is a key step towards having risk-aware robots, which can reason about the probability of successfully completing a task and act accordingly to reduce the probability of task failures.

There exists a body of work on deriving analytical solutions to the problem of uncertainty quantification for specific perception algorithms. The resultant solutions, however, often provide weak lower bounds for the uncertainty in the output of perception~\citep{kaess2009covariance}.
Solving perception problems using non-parametric implementations of Bayesian inference such as Gaussian Processes provides an estimate on the full distribution of the output of perception~\citep{urtasun20063d}; nevertheless, these approaches are computationally too expensive to solve complex real-world perception problems.

We present introspective perception, a novel approach for predicting accurate estimates of the uncertainty of perception algorithms deployed on mobile robots.
Introspective perception leverages the redundancy in the data collected by robots and enforces spatio-temporal consistency constraints across different sensing modalities to evaluate the errors in the output of perception in the deployment environment and without human supervision. 
It then uses the autonomously labeled data to learn to predict the error distribution of the output of perception as a function of the raw sensory data. 
Unlike state-of-the-art uncertainty estimation methods, introspective perception continually improves its accuracy in the deployment environment in an autonomously supervised manner.

In this paper we present a general theory for introspective perception and demonstrate successful implementations for two different perception algorithms: visual simultaneous localization and mapping and stereo depth estimation. 
We present empirical results on real-robot data to demonstrate that introspective perception learns to accurately predict errors in the output of perception without requiring human labeled data and leverages this information to reduce state estimation failures for an autonomous mobile robot.
This paper extends our prior works on developing introspection capabilities for specific perception algorithms~\citep{rabiee2019ivoa, rabiee2020ivslam}. In this work, we present a novel overarching theory for introspective perception that provides a guideline for how to enable arbitrary perception algorithms with introspection. Moreover, we for the first time implement and evaluate introspective perception for the case where the perception algorithm is an end-to-end machine learning algorithm. 

\section{Uncertainty in Perception}
Uncertainties in the output of perception stem from different sources such as shortcomings of the underlying models, partial observability, measurement noise, and etc. These uncertainties are usually classified in two main categories given their source:
\begin{inparaenum}[1) ]
\item epistemic uncertainty, and
\item aleatoric uncertainty.
\end{inparaenum}

\subsection{Epistemic Uncertainty}
Epistemic uncertainty, also known as model uncertainty, is due to the inability of the model of perception to explain a phenomena in the general case. 
This form of uncertainty happens when the sensory input is sampled from part of the observation space, where assumptions made by the perception model do not hold, or in the case of model-free perception, it happens when the input belongs to part of the observation space which is considered out of distribution (OOD) for the training data.
Examples for this are the uncertainty of an stereo reconstruction algorithm in estimating the depth of texture-less objects, or the uncertainty of an image segmentation neural network that is only trained on outdoor scenes in segmenting an indoor image.

\subsection{Aleatoric Uncertainty}
The second form of uncertainty, which is due to the noise in the data is known as aleatoric uncertainty.
This type of uncertainty happens when there is noise in the sensory input or when there exists partial observability. 
Examples of this include the uncertainty of an object classifier in discerning different but similar looking types of flowers from each other or the uncertainty of a machine learned depth estimator in predicting the depth of a shadowed part of the scene.

\section{Methods for Estimating Perception Uncertainty}
Quantifying uncertainty of the output of perception has long been an active area of research due to the great applications that it empowers. 
A perception algorithm that provides accurate confidence intervals or error bars for its output, enables the autonomous agent to do competence-aware planning~\citep{pereira2013risk}, leverage this information to improve perception via active learning~\citep{kapoor2007active}, or even detect adversarial attacks~\citep{liu2019universal}.

\subsection{Uncertainty Estimation for Model-Based Perception}

In model-based perception algorithms, it is common to use an unbiased estimator to determine the value of a set of desired parameters from a number of observations. This allows us to use the Cramer-Rao Lower Bound (CRLB) to estimate a lower-bound on the aleatoric uncertainty in the form of the covariance of the parameters of interest. This approach is common in various perception tasks such as simultaneous localization and mapping (SLAM)~\citep{kaess2009covariance}, calibration~\citep{pandey2015automatic}, and tracking~\citep{zhang2005dynamic}. The problem with CRLB is that it is usually a weak bound. Moreover, it is computed conditioned on accurate data association, and assumes accurate mathematical models of the observation likelihood functions. Invalidation of these assumptions in the real-world applications such as robotics reduce the accuracy of CRLB.

In order to also address the epistemic uncertainty due to the shortcomings of the model itself and achieve a tighter bound on the total uncertainty of perception,
one approach is to leverage expert knowledge about the internals of a perception algorithm to create heuristic measures of its reliability. Examples of this include using the density of image features or the viewing angle of map points at every frame to calculate a hand-crafted measure of the reliability of visual SLAM algorithms~\citep{sadat2014feature, mostegel2014active}. Similar to any hand-crafted heuristic these measures of uncertainty fail to cover all potential sources of failure.

\subsection{Uncertainty Estimation for Model-Free Perception}\label{sec:uncertainty_est_model_free}

Using Bayesian inference for non-parametric and model-free perception is a principled way for quantifying the total uncertainty, i.e. the sum of aleatoric and epistemic uncertainty, of such perception algorithms. There exist exact and approximate Bayesian inference methods.

\subsubsection{Bayesian Inference}
One of the classical examples of Bayesian inference is using Gaussian Processes (GP) for regression or classification tasks~\citep{rasmussen2003gaussian, urtasun20063d}. GP regression (GPR) is a specific form of Bayesian inference where the prior is a multivariate Gaussian distribution and the goal is to calculate the probability distribution over the space of functions that fit the training data as opposed to the probability distribution of parameters of a specific model.
Using the Gaussian process prior and a Gaussian likelihood allows us to analytically derive a closed form solution to the posterior distribution in the Bayes formulation.
However, the complexity of computing the posterior is still cubic with respect to the number of training data. 
In order to improve the scalability, dimensionality reduction techniques are applied to the high dimensional input data and only a subset of the training data is used to compute the predictive posterior distribution~\citep{quinonero2005unifying}.
This in turn limits the applicability of such methods for complex perception problems with high dimensional input data.

Deep learning models have shown great success in real-world perception tasks due to their power in automatic feature extraction from high-dimensional data. However, these methods do not come with inherent uncertainty estimation capabilities.
Recently Bayesian neural networks (BNN)~\cite{ghahramani1997learning} were introduced as a method to realize Bayesian inference with these models. Compared to the standard NNs where each parameter in the network is a single point estimate, in BNN each parameter is represented as a probability distribution from which we draw samples, and these distributions evolve away from their prior during training. The output of a BNN is a probability distribution that includes information about the total uncertainty. It has been proven that a single-layer infinitely wide BNN is equivalent to a Gaussian process~\cite{lee2017deep}.  
While promising results have been achieved using BNNs on small problems~\cite{dusenberry2020efficient}, the main issue with this method, which has prevented it from being used in real-world perception applications is that it is slow to train and difficult to scale. 

While obtaining estimates of the full probability distribution of outputs given inputs and model parameters is ideal, implementation of Bayesian inference for complex real-world robotic applications is intractable. Both training and inference on such models are computationally expensive. An alternative approach is using approximation methods for Bayesian inference.

\subsubsection{Approximate Bayesian Inference}
In order to address the computational cost of BNN, approximation methods have been proposed. \citet{gal2016dropout} showed that using Monte Carlo (MC) dropouts in a NN during inference is a valid approximation of a BNN. In this approach, some layers of a deterministic NN are equipped with dropout units that are used during inference as well as training and their job is to drop a node conditioned on the outcome of sampling from a Bernouli distribution. Each input is passed through the network several times, and the output of all these different trials are used to represent an approximate distribution of the output. 
Using dropouts during inference can be interpreted as averaging many models with shared weights, where each model is equally weighted. 
This is different from BNN, where each model is weighted taking into account priors and how well the model fits data. 

Averaging the output of several models is also the main idea behind ensemble techniques for estimating the uncertainty of machine learned models. Ensemble methods have been applied to both classical models such as SVMs~\cite{grimmett2016introspective} as well as more recently for deep learning~\citep{lakshminarayanan2017simple}. In the case of deep ensembles the same NN architecture is re-trained several times with different random seeds and the output of the collection of these models for the same input is used to approximate the uncertainty of the output. 
This technique relies on different initializations of the network parameters as well as the noise in stochastic gradient descent (SGD) for different instances of the same model to converge to different local optima after training.
It has been shown that the approach can capture both aleatoric and epistemic uncertainty to some extent~\citep{ovadia2019can}. The problem with both MC dropout and ensemble methods is that their run-time and required memory linearly increase with respect to the number of models in the ensemble or the number of drawn samples. This is specifically more pronounced with large perception models that tackle complex real-world problems.

\subsection{Out-of-Distribution and Anomaly Detection}
There exists a large body of work on out-of-distribution (OOD) detection, where the goal is to detect whether a given input is from the same distribution as the training data. 
Some of the popular methods for addressing this problem are generative adversarial networks (GANs)~\citep{goodfellow2014generative}, autoencoders (AE)~\citep{vincent2008extracting}, and variational autoencoders (VAEs)~\citep{kingma2013auto, an2015variational}. The common idea behind these methods is to learn an embedding function that maps the data from the training set to a hyperspace, such that mappings of new OOD data points fall outside of this hyperspace.
In the case of AEs, an example input is detected to be OOD if the reconstruction error is high. In the case of VAEs,
where the latent space is assumed to be Gaussian, a novelty score is computed as the KL divergence between the latent space distribution of the input and the prior distribution.

While these methods are effective in detecting OOD data, which can potentially lead to high epistemic uncertainty, they cannot capture aleatoric uncertainty of perception.
Nevertheless, making accurate predictions of the aleatoric uncertainty of perception and accounting for that in the planning is especially important since this type of uncertainty cannot be reduced by improving the perception function.

\changes{
\subsection{Formal Verification Techniques for Runtime Monitoring of Perception Systems}
There is a body of research focused on utilizing formal verification techniques~\citep{desai2017combining, ghosh2016diagnosis} to analyze the behavior and ensure the safety of mobile robots at runtime. These techniques use formal specifications, such as temporal logic~\citep{balakrishnan2021percemon} to define desired properties of the perception algorithms. By monitoring the outputs of the perception algorithms and comparing them with the formal specifications, deviations and anomalies can be detected.
Formal verification techniques often rely on high-level abstractions of the perception algorithms~\citep{balakrishnan2019specifying} and require accurate specifications that align with the specific characteristics of the perception algorithms being tested~\citep{mitsch2017formal}. Defining these specifications can be difficult and cumbersome in practice, especially for complex perception algorithms.
}

\subsection{Learning to Predict Perception Errors}
The available analytical solutions for uncertainty quantification in the case of model-based perception algorithms, e.g. information-theoretic metrics such as the Fisher information metric to compute lower bounds for the uncertainty of perception~\citep{kaess2009covariance}, are conditioned on accurate data association, hence provide overconfident estimates of perception uncertainty in real-world applications.
Also, for end-to-end machine learning perception algorithms, the available uncertainty estimation methods are either computationally expensive or are application-specific and require modifications to the architecture of the network if applied to off-the-shelf perception models.
A different approach that has recently gained interest includes learning a secondary machine-learned model to predict the probability of failure of the perception algorithm at hand. 
This technique, referred to as introspective perception, aims to provide a measure of reliability for the perception algorithm.
The failure prediction model does not require any information about the underlying details of the perception algorithm. Instead, it relies on raw sensory data to predict the probability of failure in perception.
Zhang et al.~\cite{zhang2014predicting} use labeled data and train a binary classifier, which given an input image predicts the success or failure of a vision system. They test their method for the two different tasks of image classification and image segmentation. Daftry et al.~\cite{daftry2016introspective} train a convolutional neural network (CNN) that uses both still images and optical flow frames to predict the probability of failure for obstacle detection on an actual UAV.
\citet{guruau2018learn} develop an environment-specific failure prediction model that uses the raw sensory data as well as the location information of the robot to predict the probability of failure of a perception algorithm given examples of its prior failures in the same environment.
This body of work solves the binary classification problem of failure prediction as opposed to approximating the distribution of perception errors. Moreover, the aforementioned works require the availability of sufficient labeled data of instances of perception failures for training prior to deployment.
\citet{vega2013cello} propose a method for learning to predict the covariance of state estimation errors given hand-engineered features extracted from the raw sensory data. However, they do not address the problem of collecting the training data for the covariance estimation model.

\section{Introspective Perception}

We present introspective perception (\IPrName{}), a method to empirically learn the error distribution of the output of perception in the deployment environment of a robot.
Introspective perception provides the means to accurately estimate the uncertainty in the output of perception algorithms in a computationally efficient manner.
\IPrName{} applies to arbitrary perception algorithms and learns the distribution of error in their output as a function of the raw sensory data.
The key property of \IPrName{} that makes the learning of the error distribution scalable is that it is autonomously supervised. 
Inspired by prior works that leverage the redundancy of information on robots to perform self-calibration \cite{stronger2006towards, holtz2017deltacalibration}, \IPrName{} also exploits this resource, but to learn to predict the uncertainty of perception.
It collects the training data required for learning the error distribution by enforcing consistency constraints on the outputs of perception algorithms. 
This allows \IPrName{} to leverage the abundance of data logged by mobile robots in the deployment environment to improve the accuracy of its estimate of the uncertainty of perception.
In this section we present the mathematical definition of introspective perception and explain the autonomously supervised training techniques used by \IPrName{}.

\subsection{Learning to Predict Errors of Perception}

\begin{table}[h]  
\caption{\textsc{Table of notations.}}
\label{tab:notations}
\centering
\resizebox{0.7\linewidth}{!}{%
\begin{tabular}{@{}ll@{}}
\toprule
Notation                     & Description                                                                  \\ \midrule
$\perc(\cdot)$                   & Perception function                                                          \\
$\stateWorldSet$            & Set of world states \\
$\stateWorld{t}$            & State of the world at time $t$ \\
$\stateOutSet{\perc}$       & Set of world states estimated by $\perc$ \\
$\stateOut{\perc}{t}$       & Ground truth state at time $t$ of states estimated by $\perc$              \\
$\stateOutEst{\perc}{t}$    & The estimates from $\perc$ at time $t$ \\

$\stateInSet{\perc}$     & Set of world states required by $\perc$          \\
$\stateIn{\perc}{t}$        & Ground truth state at time $t$ of states required by $\perc$   \\
$\stateInEst{\perc}{t}$     & The estimates at time $t$ of states required by $\perc$    \\
$\obsAllSet$                & Set of all observations (sensor readings) by the robot \\
$\obsPercSet{\perc}$        & Set of all observations processed by $\perc$ \\
$\obsPerc{t}{\perc}$        & Observations at time $t$ processed by $\perc$ \\
$\errDist$                  & PDF of the error distribution of the output of F                                        \\
$\transform{i}{j}{\cdot}$     & Transformation of the world state from time $i$ to time $j$.                   \\ \bottomrule
\end{tabular}
}
\end{table}

In order to motivate our work, we formulate a perception algorithm as a function that produces estimates of a subset of the world states given a set of raw sensory observations as well as prior estimates of a subset of the world states
\begin{align}
\percFunc{\cdot} &: \obsPercSet{\perc} \times \stateInSet{\perc} \rightarrow \stateOutSet{\perc} \\
\stateOutEst{\perc}{t} &= \percFunc{\obsPerc{t}{\perc}, \stateInEst{\perc}{t-1}}
 = \percFunc{\obsPerc{t}{\perc}, \left<  \stateOutEst{\perc_0}{t-1}, \stateOutEst{\perc_1}{t-1}, ..., \stateOutEst{\perc_k}{t-1} \right>}    \\
\stateOutEst{\perc}{t} &\in \stateOutSet{\perc} \subseteq \stateWorldSet 
\end{align}
where $\obsPerc{t}{\perc}$ are raw sensor measurements while $\stateInEst{\perc}{t}$ are the intermediate state factors that are the result of processing the raw sensory data by other perception algorithms $\perc_{0}, ..., \perc_{k}$. Table~\ref{tab:notations} lists the description of all the parameters and variables used in the problem formulation.
Consider the example of a human tracking algorithm $\perc{}$. Then at any time-step $t$, $\obsPerc{t}{}$ is the image captured by an RGB camera and $\stateInEst{\perc}{t-1}$ is a tuple of the bounding boxes of pedestrians in the image as detected by an object classifier $\perc^{\prime}$ along with the previous time-step estimate of the 3D pose of the pedestrians $\stateOutEst{\perc}{t-1}$ by $\perc{}$.
The human tracking algorithm $\perc{}$ would use $\obsPerc{t}{}$ and $\stateInEst{\perc}{t-1}$ to estimate the pose of the pedestrians at the current time step $\stateOutEst{\perc}{t}$.

The perception error can then be written as 
\begin{align}
\stateOutEst{\perc}{t} &= \stateOut{\perc}{t} \oplus \err, \qquad \err \sim \errPDFDep \, .
\end{align}
Note that $\errPDF$ is non-stationary and depends on the observations and the state of the agent. The fact that the perception error distribution depends on factors other than only the parameters of the perception function makes it challenging to derive analytical noise models that accurately estimate the uncertainty of the perception output. To address this, introspective perception learns an empirical approximation of the error distribution.

\begin{figure*}[t]
  \centering 
  \includegraphics[width=0.6\linewidth ,trim=0 0 0 0,clip]{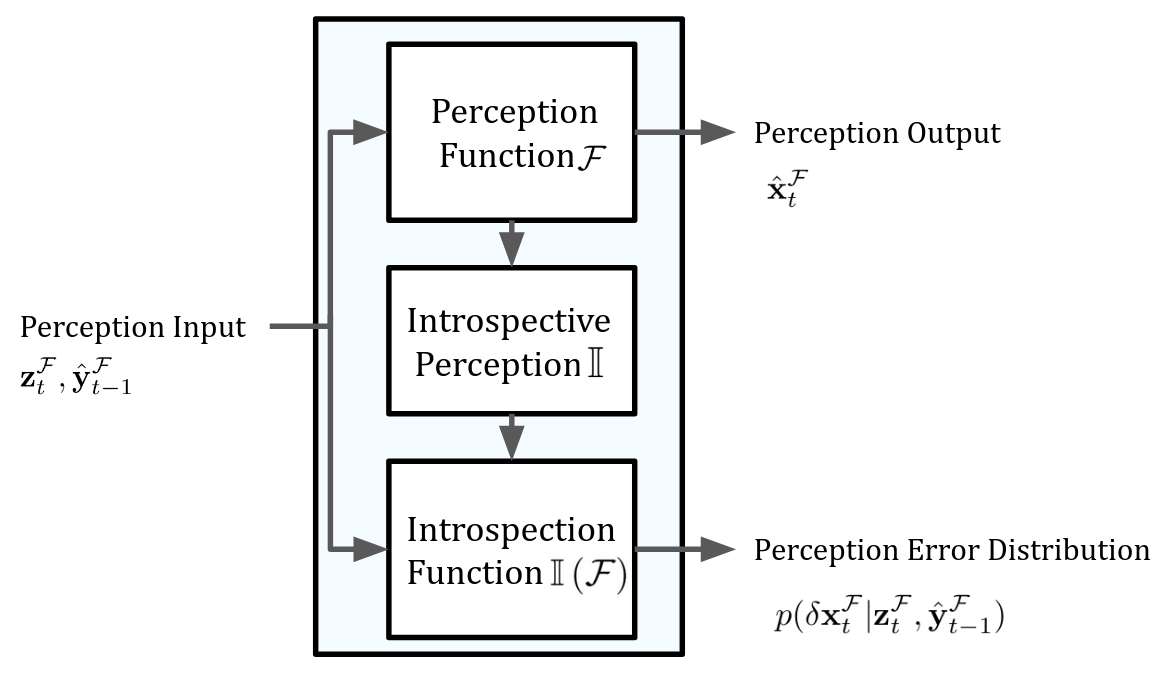}
  \caption{Architecture of introspective perception. Introspective perception is a higher-order function $\IPrShortest$ that when applied to a perception function $\perc$, produces a function that predicts the error distribution of $\perc$ conditioned on the inputs.}
  \label{fig:ipr_architecture}
\end{figure*}

We define introspective perception as a higher-order function $\IPrShortest$ that when applied to a perception function $\perc$, produces a function that predicts the error distribution of $\perc$ for given inputs
\begin{align}
\errPDFEst &= \IPr \, ,
\label{eq:ipr_def}
\end{align}
where the resultant function $\IF{\perc}: \obsPercSet{\perc} \times \stateInSet{\perc} \rightarrow \stateOutSet{\perc}$ is the \emph{introspection function}. Figure~\ref{fig:ipr_architecture} illustrates the architecture of introspective perception. \changes{It should be noted that not all perception functions necessarily include the intermediate state factors $\stateInEst{\perc}{}$ as their input. In this work, we adopt a more general definition to demonstrate the applicability of introspective perception across a broader range of perception functions, but $\stateInEst{\perc}{}$ can be omitted from the formulations for the specific case of perception functions that solely rely on raw sensor data $\obsPerc{t}{\perc}$ as their input.}

\subsection{Autonomously Supervised Training of Introspective Perception}\label{sec:ipr_training_method}

In order to learn the introspection function $\IF{\perc}$, we need training data in the form of recorded errors of the output of $\perc$ for various input values of $\obsPerc{t}{\perc}$ and $\stateInEst{\perc}{t-1}$. Manual labeling of the required training data is cumbersome and the resultant learned introspection function will 
only be accurate in environments for which we have labeled data.
Mobile robots, on the other hand, collect a great amount of raw data during deployment. The collected data, although unlabeled, includes redundancy across different sensing modalities and across different time steps. 
We leverage the naturally available redundancy in data to generate the required labeled data autonomously.
Specifically, we propose the following two consistency constraints to autonomously generate the training data for the introspection function: 1) consistency across sensors, and 2) spatio-temporal consistency.

\paragraph{Consistency Across Sensors}
Let $\stateOut{\perc}{t} \approx \stateOutEst{\perc'}{t}$, where $\perc'$ is a supervisory perception function that consumes observations $\obsPerc{t}{\perc'}$ from high-fidelity supervisory sensors. Then 
\begin{align}
\stateOutEst{\perc}{t} \ominus \stateOutEst{\perc'}{t} \sim \errPDFDep \quad .
\label{eq:sensing_consitency}
\end{align}
In other terms, we can autonomously collect samples of errors of a perception function $\perc$ if we have access to a high-fidelity supervisory sensor on our robot that can be used by another perception function $\perc'$ to provide accurate estimates of the world state.
When deploying a fleet of robots in a real-world environment, a cost-effective way to achieve this is to equip only some of them with additional high-fidelity sensors. 
For example, most of the robots in the fleet might use stereo RGB cameras for depth estimation while a few of them can additionally be equipped with more expensive long-range and high-resolution Lidars to obtain high-fidelity depth estimations of the environment and use them to provide supervision for passive depth estimation using stereo RGB cameras. 
Moreover, the supervisory perception function $\perc'$ may be computationally expensive and not executable in real time; however, it can be used offline for supervision.
For example, $\perc'$ can be a structure from motion (SfM) algorithm~\citep{schonberger2016structure} for dense 3D reconstruction of the environment or a depth estimator that is based on Neural Radiance Fields (NeRF)~\citep{kundu2022panoptic}. These algorithms can be used offline to process the recorded data by the robot and provide supervision for the primary and real-time perception algorithm $\perc$.

\paragraph{Spatio-temporal Consistency}
By leveraging spatio-temporal constraints, we can relate state variables at different time-steps such that
\begin{align}
\transform{i}{j}{\cdot} &: \stateOutSet{\perc} \rightarrow \stateOutSet{\perc} \\
\transform{i}{j}{\stateOut{\perc}{i}} &= \stateOut{\perc}{j} \label{eq:spatio_temporal_constraint} \\
\stateOutEst{\perc}{t} &= \transform{t-\delta t}{t}{\stateOut{\perc}{t- \delta t}} \oplus \err \quad ,
\end{align}
where $\transform{i}{j}{\cdot}$ is the transformation function converting the world state from time $i$ to time $j$.
For an incremental perception algorithm, it is reasonable to assume that the posterior estimate of a state variable $\stateOutEstPosterior{\perc}{t - \delta t}{t}$, i.e. the estimate of $\stateOutEst{\perc}{t - \delta t}$ made at time $t$ improves in accuracy as $\delta t$ increases. Hence, assuming $\stateOut{\perc}{t - \delta t} \approx \stateOutEstPosterior{\perc}{t - \delta t}{t}$, we have 

\begin{align}
\stateOutEst{\perc}{t} &\approx \transform{t-\delta t}{t}{\stateOutEstPosterior{\perc}{t- \delta t}{t}} \oplus \err \\
&\Rightarrow 
\stateOutEst{\perc}{t} \ominus \transform{t-\delta t}{t}{\stateOutEstPosterior{\perc}{t-\delta t}{t}} \sim \errPDFDep
\label{eq:spatio_temporal_consitency}
\end{align}

\paragraph{Training}
Consistency across sensors and spatio-temporal consistency as shown in Eq.~\ref{eq:sensing_consitency} and Eq.~\ref{eq:spatio_temporal_consitency} provide the means for generating samples of perception errors without the need for manual inspection and labeling of the data. 
Let $\errSampleSet = \lbrace \errSample{1}, \errSample{2}, ..., \errSample{N} \rbrace$ be the empirical distribution of errors of $\perc$, where $\errSample{i} \sim \errPDF$ are samples of errors of the output of $\perc$ that are obtained using spatio-temporal consistency constraints as well as consistency across sensors. 
\IPrName{} learns an approximation $\IPrShort$ of the introspection function such that 
\begin{align}
\theta^* = \argmin_{\theta} \KLD\left( \IPrTheta , \errSampleSet \right) \quad ,
\label{eq:introspection_func_training}
\end{align}
where $\KLD$ is the Kullback-Leibler divergence.

The introspection function $\IFShortTheta{\perc}$ can accurately predict the error distribution of $\perc$ when its input data is from the same distribution as the training data of $\IFShortTheta{\perc}$, i.e. $\errSampleSet$. 
While similar to any machine learning model, $\IFShortTheta{\perc}$ is prone to errors when its input data is out of distribution (OOD) of the training data, the autonomously supervised training of introspective perception ensures that the occurrence of OOD observations is minimized. 
\IPrName{} collects training data from all deployment environments of the robots at a low cost and without a human in the loop.

It should be noted that the solution to the optimization problem in Eq.~\ref{eq:introspection_func_training} changes if modifications are made to the perception function $\perc$, e.g. if the perception function is re-trained on new data. As a result, the introspection function needs to be re-trained when the perception function is updated. However, a reasonable assumption is that incremental updates to the perception function do not change the error characteristics of the perception function significantly; therefore, the introspection function can be fine-tuned from its latest state as opposed to re-trained from scratch, which in turn reduces the computational overhead of updating the introspection function.

\subsection{Design and Implementation of the Introspection Function}
While in theory, the introspection function can be implemented as any function approximator, we use deep learning models due to their strength in learning generalizable representations of their input data. The choice of network architecture can vary for different perception tasks. A good network architecture depends on the type of sensor data consumed by the perception algorithm, the difficulty of the learning task, and the size of the training data set that can be collected using the available consistency constraints.
In the design and implementation of the network architecture for each new perception task, we take into account three main points. 1) We use encoder architectures that have been shown to successfully learn representations of the type of sensor data consumed by the perception function. For instance, convolutional neural networks are a great choice for learning representations of image data while recurrent neural networks are a great choice for learning representations of sequential data such as IMU readings.
2) We keep the network architecture simple and its number of parameters small to reduce the chance of overfitting the training data and to ensure that the network can be trained efficiently and can be deployed in real time.
3) We parametrize the error distribution $\errPDF$ which is the output of the introspection function such that it simplifies the learning task. For instance, instead of having the introspection function estimate the distribution of the error vectors, we task the introspection function to estimate the distribution of the magnitude of the error vectors. We introduce example architectures for different perception tasks in sections~\cref{sec:ivslam} and ~\cref{sec:ivoa}.

\section{Improved Robustness in Autonomy via Introspective Perception}
Equipping perception algorithms with introspection improves robustness and reliability of an autonomous robot in two main ways. First, it improves accuracy of perception algorithms by producing more accurate uncertainty estimates for intermediate state variables that are used by high-level perception algorithms. Second, uncertainty estimates produced by introspective perception, can be leveraged in planning to prevent task-level failures. In this section we explain each of these two benefits of introspective perception in detail.

\subsection{Reducing Errors of Perception}

Perception algorithms in the general case keep a belief over state variables of interest
\begin{align}
\bel \left( \stateOut{\perc}{t} \right) &= 
p(\stateOut{\perc}{t} | \obsPerc{t}{\perc}, \stateInEst{\perc}{t-1}) \propto  
p( \stateInEst{\perc}{t-1} | \stateOut{\perc}{t}, \obsPerc{t}{\perc} ) 
p(\obsPerc{t}{\perc} | \stateOut{\perc}{t}) p(\stateOut{\perc}{t}) \\
 &\propto  p( \stateInEst{\perc}{t-1} | \stateOut{\perc}{t}) p(\obsPerc{t}{\perc} | \stateOut{\perc}{t}) p(\stateOut{\perc}{t}) \quad .
\end{align}
A large subset of perception algorithms, however, compute a maximum likelihood estimate (MLE) of the state variables 
\begin{align}
\stateOutEst{\perc}{t} 
 &= \argmax_{\stateOut{\perc}{t}} p( \stateInEst{\perc}{t-1} | \stateOut{\perc}{t}) p(\obsPerc{t}{\perc} | \stateOut{\perc}{t}) \quad .
\end{align}
The way different perception algorithms solve this problem varies given the task at hand. 
There are low-level perception algorithms for which the there exist a unique solution to the above problem given the number of available observations and the degree of freedom of the problem. An example for this is depth estimation for associated pairs of image features in stereo images, which can be solved analytically.
However, it is often the case that the problem solved by $\perc$ is over-determined hence obtaining the output involves solving an optimization problem which minimizes a loss function
$\mathcal{L}(\stateOut{\perc}{t}) = -\log p(\stateInEst{\perc}{t-1} | \stateOut{\perc}{t}) -\log p(\obsPerc{t}{\perc} | \stateOut{\perc}{t}) $.
The term
$p(\obsPerc{t}{\perc} | \stateOut{\perc}{t})$ captures the noise distribution of the sensor readings and is obtained via sensor-specific calibration processes. An example for this is the precision of LiDAR range measurements given the surface material of obstacles or the accuracy of sonar range finders given the humidity of the environment.
The conditional probability
$p( \stateInEst{\perc}{t-1} | \stateOut{\perc}{t})$ captures both the epistemic uncertainty in the perception algorithm $\perc^{\prime}$ that estimates $\stateInEst{\perc}{t-1}$ as well as the aleatoric uncertainty in the data processed by $\perc^{\prime}$.
In the absence of an accurate estimate of $p(\stateInEst{\perc}{t-1} | \stateOut{\perc}{t})$, it is often approximated to be static and from one of the known families of distributions, e.g. Gaussian. 
However, this is a poor approximation of the true distribution of the error which is often heteroscedastic.

Note that $\stateInEst{\perc}{}$ is itself the output of another perception algorithm $\perc^{\prime}$, i.e.
$\stateInEst{\perc}{t} = \stateOutEst{\perc^{\prime}}{t}$. Therefore, if we 
equip $\perc^{\prime}$ with introspection such that $\errPDFEstPerc{\perc^{\prime}} = \IPrPerc{\perc^{\prime}} $, 
we can provide $\perc$ with the $\stateInEst{\perc}{}$ error distribution as predicted by the $\perc^{\prime}$ introspection function. Leveraging the resultant more accurate approximation of $p(\stateInEst{\perc}{t-1} | \stateOut{\perc}{t})$ in the loss function $\mathcal{L}(\stateOut{\perc}{t})$ leads to a more accurate estimate of $\stateOut{\perc}{t}$.
The same technique also applies to perception algorithms that output a belief over the state variables, e.g. particle filter-based localization or tracking. Using a learned and more accurate approximation of the observation likelihood $p(\stateInEst{\perc}{t-1} | \stateOut{\perc}{t})$ leads to a better estimate of $\bel \left( \stateOut{\perc}{t} \right)$.

\subsection{Preventing Task-Level Failures}
The approximation of distribution of error for the state variables, provided by introspective perception, can also be leveraged in planning to prevent task-level failures that arise from errors in perception, e.g. collision with obstacles due to depth estimation or localization errors.

There exist several planning methods that take into account the uncertainty in the state of the agent when planning.
Partially Observable Markov Decision Processes (POMDP) provide the means for computing an optimal policy in the belief space, i.e. when the agent keeps a distribution of states~\citep{thrun2005probabilistic}. POMDPs use an observation likelihood function to update their belief over the state of the world given new observations, hence, a more accurate approximation of the observation likelihood function helps improve their performance.
The main down-side of POMDPs is their high computational cost. More tractable solutions include set-bounded uncertainty modeling, where it is assumed that a bound on the magnitude of perception and actuation error is known and then planning is limited to a subset of the world states where safety of the plan is guaranteed as long as the error values are within the assumed bounds~\citep{pepy2006safe, ostafew2016robust}. 
Chance-constrained optimization methods relax the assumption of requiring bounds on the magnitude of perception and actuation errors, and leverage a known error distribution to solve the planning problem subject to constraints on the probability of failures~\citep{barbosa2021risk,blackmore2011chance,jasour2019risk}. 
While the above approaches consume continous perception error probability distributions, there also exist learning-based risk-aware planning algorithms that require only discrete and sparse histogram representations of the error distributions of low-level perception to predict the probability of failure at the task level, e.g. autonomous navigation~\citep{rabiee2021competenceaware}.

The effectiveness of all above planning methods depends on having accurate models of the uncertainty of perception, and introspective perception provides the means to achieve this in the deployment environment of the robot and in a computationally efficient manner.

\section{Introspective Perception vs. Approximate Bayesian Inference}\label{sec:approximate_BNN}

As explained in~\cref{sec:uncertainty_est_model_free}, sampling-based approximate Bayesian inference methods provide the means to estimate the uncertainty of a function approximator in a data-driven manner that is scalable in terms of the number of training data samples. In this section, we explain approximate Bayesian inference in detail and discuss how it compares to introspective perception.

For a perception function approximated by $\perc_{\theta}\left(.\right)$, where $\theta$ is a random vector and the parameters of the function, approximate Bayesian inference generates a set of functions
\begin{align}
\lbrace \perc_{\theta^{(i)}}() \rbrace_{i = 1}^{N} &: \theta^{(i)} \sim \Theta \quad , \end{align}
where $\theta^{(i)}$ is a sample from the distribution of parameters $\Theta$. Then for each input pair $\left(\obsPerc{t}{\perc}, \stateInEst{\perc}{t-1}\right)$, the output of all $\perc_{\theta^{(i)}}$ are computed to generate an empirical distribution of the output of $\perc_{\theta}$ at time $t$. For instance the expected value of the output is estimated as
\begin{align}
\hat{\bar{\mathbf{x}}}_{t}^{\perc} &= \frac{1}{N} \sum_{i=1}^{N}  \perc_{\theta^{(i)}}(\obsPerc{t}{\perc}, \stateInEst{\perc}{t-1}) \quad .
\end{align}
Assuming $\perc_{\theta}$ is an unbiased estimator, then $\hat{\phi}^{\perc} = \lbrace \perc_{\theta^{(i)}}(\obsPerc{t}{\perc}, \stateInEst{\perc}{t-1}) -  \hat{\bar{\mathbf{x}}}_{t}^{\perc} \rbrace_{i = 1}^{N}$ forms an empirical distribution approximating the true distribution of error of $\perc_{\theta}$ at time $t$.

For the case when $\perc_{\theta}()$ is implemented as a neural network, computationally tractable approximate Bayesian inference methods have been proposed. The most popular of these methods are the Monte Carlo Dropout (MCDropout) and deep ensembles. In the case of MCDropout, each $\perc_{\theta^{(i)}}$ is a copy of the network with the resultant weights after a random sampling of dropout units in the network. In the ensemble approach, each $\perc_{\theta^{(i)}}$ is the result of a different round of training the network with a different initialization of the weights and different sampling of the training data.
In both approaches, the ensemble of $\perc_{\theta^{(i)}}$ is treated as a uniformly-weighted mixture model when combining the predictions. In the case of the perception function being a classifier, combining the results would be equivalent to averaging the predicted probability values for each class. In the case of regression, it is common to treat the prediction as a mixture of Gaussian distributions such that $\perc_{\theta^{(i)}}(\obsPerc{t}{\perc}, \stateInEst{\perc}{t-1}) \sim \mathcal{N}(\mu_{\theta^{(i)}}\left(\obsPerc{t}{\perc}, \stateInEst{\perc}{t-1}  \right), \sigma^2)$, where $\sigma^2$ is the variance of the output of each model and defined by the user as a hyperparameter.
The ensemble prediction is then approximated to be a Gaussian with its mean and variance given by
\begin{align}
\mu_{\text{ens}}(\obsPerc{t}{\perc}, \stateInEst{\perc}{t-1}) &= \frac{1}{N} \sum_{i=1}^{N} \mu_{\theta^{(i)}}\left(\obsPerc{t}{\perc}, \stateInEst{\perc}{t-1}  \right) \\
\sigma_{\text{ens}}^2(\obsPerc{t}{\perc}, \stateInEst{\perc}{t-1}) &= \frac{1}{N} \sum_{i=1}^{N} \left( \sigma^2 + \mu_{\theta^{(i)}}^2\left(\obsPerc{t}{\perc}, \stateInEst{\perc}{t-1}  \right)  \right) - \mu_{\text{ens}}^2(\obsPerc{t}{\perc}, \stateInEst{\perc}{t-1}) \quad .
\end{align}
Compared to \IPrName{}, approximate Bayesian neural networks (BNN) also use a machine learned function approximator to do uncertainty estimation.
For BNNs, the machine learned component used for uncertainty estimation is the same as that used for the perception function. 
\IPrName{}, however, is agnostic of the underlying function approximator used for the perception function. 
The machine learned component that it uses for uncertainty estimation, i.e. the introspection function, is different from the perception function and specifically trained for estimating the error distribution of perception. Therefore, \IPrName{} applies to both model-based and fully-learned perception algorithms.
Additionally, the modular architecture of \IPrName{} makes it computationally more efficient. While BNNs' usage of memory and inference time scales linearly with the number of Monte Carlo samples or the models in the ensemble, \IPrName{} requires a constant amount of memory to run the introspection function. The introspection function can be implemented with a significantly less complex function approximator than the perception function since the task tackled by the introspection function is simpler than the full perception problem. As a result, the computational requirements of \IPrName{} are significantly less than those of BNNs.

In~\cref{sec:result_ipr_fully_learned_perc}, we discuss the results of using \IPrName{} for a fully-learned perception algorithm and compare its performance in terms of perception failure prediction accuracy and also its computational cost with approximate Bayesian inference methods.

\section{Introspective Visual Simultaneous Localization and Mapping}\label{sec:ivslam}
In this section we present an implementation of introspective perception for Visual SLAM and demonstrate how spatio-temporal consistency constraints can be used to train an introspection function which in turn can be leveraged to reduce failures of V-SLAM.

\subsection{Visual SLAM}
In visual SLAM, the pose of the camera $\camPose \in SE(3)$ is estimated in the world reference frame $w$ and a 3D map $\mathbf{M} = \left\lbrace \mathbf{p}_k^w | \mathbf{p}_k^w \in \mathbb{R}^{3} , k \in [1, N]   \right\rbrace$ of the environment is built by finding correspondences in the image space across frames. 
The V-SLAM architecture, with the exception of the end-to-end learning approaches~\citep{wang2017deepvo, parisotto2018global}, consists of two main perception modules. The first module is the SLAM front-end $\slamFE$ that is responsible for feature extraction and matching across images and 
\begin{align}
\stateOutEst{\slamFE}{t} = \slamFE \left( \obsPerc{t}{\slamFE}, \stateOutEst{\slamFE}{t-1} \right) \, ,
\end{align}
where $\obsPerc{t}{\slamFE} = \img{t}$ are the input images and $\visualFeatVec{t}$ are extracted and matched image features at time $t$.
The second perception module is the SLAM back-end $\slamBE$ that estimates the pose of the camera given data correspondences provided by the front-end such that 
\begin{align}
\stateOutEst{\slamBE}{t} = \slamBE \left( \obsPerc{t}{\slamBE}, \visualFeatVec{1:t}  \right) \, , 
\end{align}
where $\obsPerc{t}{\slamBE} = u_{1:t}$ are the history of odometry and IMU measurements and $\stateOutEst{\slamBE}{t} = \stateVecSolution$.
The SLAM back-end $\slamBE$ solves for the pose of the camera and the map of the environment via MLE such that
\begin{equation}
\begin{aligned}
\stateVecSolution &= \argmax_{\stateVector} p\left(\stateVector|\visualFeatVec{1:t}, \odom{t}\right) \\
&= \argmax_{\stateVector} p\left(\visualFeatVec{1:t}|\stateVector\right) p\left(\camPoseAll | \odom{t} \right) \, ,
\label{eq:slam_form}
\end{aligned}
\end{equation}
where $p(\visualFeatVec{}|\stateVector)$ is the observation likelihood for image feature correspondences, given the estimated poses of the camera and the map $\mathbf{M}$. 
For each time-step $t$, the V-SLAM frontend processes image $\mathbf{I}_t$ to extract features 
$\stateOutEst{\slamFE}{t} = \left \lbrace \visualFeatEst{t}{k} | \visualFeatEst{t}{k} \in \mathbb{P}^{2}, k \in [1, N] \right \rbrace$, where $\visualFeatEst{t}{k}$ is an image feature associated with the 3D map point $\mathbf{p}_k^w$.
The observation error here is the reprojection error of $\mathbf{p}_k^w$ onto the image $\mathbf{I}_t$ and is defined as
\begin{equation}
\reprojErr{t}{k} = \visualFeatEst{t}{k} - \visualFeat{t}{k}, \quad \visualFeat{t}{k} = \mathbf{A} \left[ \mathbf{R}^t_w | \mathbf{t}^t_w \right] \mathbf{p}_k^w \, ,
\label{eq:reprojection_error}
\end{equation}
where $\mathbf{A}$ is the camera matrix, and $\mathbf{R}^t_w \in SO(3)$ and $\mathbf{t}^t_w \in \mathbb{R}^3$ are the rotation and translation parts of $\camPoseInv = \left({\camPose}\right)^{-1}$, respectively.
In the absence of control commands and IMU/odometry measurements, Eq.~\ref{eq:slam_form} reduces to the bundle adjustment (BA) problem, which is formulated as a nonlinear optimization:
\begin{equation}
\stateVecSolution = \argmin_{\stateVector}
\sum_{t, k} \mathcal{L}\left(\reprojErrShortHand_{t, k}^T \Sigma_{t,k}^{-1} \reprojErrShortHand_{t, k}\right) \, ,
\label{eq:bundle_adjustment}
\end{equation}
where $\Sigma_{t,k}$ is the covariance matrix associated with the scale at which a feature has been extracted, $\reprojErrShortHand_{t, k} = \reprojErr{t}{k}$, and $\mathcal{L}$ is the loss function for $p(\visualFeatVec{}|\stateVector)$.

The choice of noise model for the observation error $\reprojErr{t}{k}$ has a significant effect on the accuracy of the maximum likelihood solution to SLAM~\citep{triggs1999bundle}. There exists a body of work on developing robust loss functions~\citep{black1996unification, barron2019general} %
that targets improving the performance of vision tasks in the presence of outliers.
The reprojection error is known to have a non-Gaussian distribution $\errDistSLAM$  in the context of V-SLAM due to the frequent outliers that happen in image correspondences~\citep{triggs1999bundle}. It is assumed to be drawn from long-tailed distributions such as piecewise densities with a central peaked inlier region and a wide outlier region. 
\changes{In addition to using robust loss functions, outlier rejection methods are used to detect wrong data associations and remove the corresponding observation error terms from the optimization problem in Eq.~\ref{eq:bundle_adjustment}. Common outlier rejection methods in V-SLAM include RANSAC~\citep{fischler1981random}, detecting outliers based on the temporal consistency of data correspondences~\citep{forster2014svo}, and learning-based methods~\citep{yi2018learning, sun2020acne} which use permutation-equivariant network architectures and predict outlier correspondences by processing coordinates of the pairs of matched features. While these methods are effective in removing gross outliers, not all bad image feature matchings are clear outliers; there is an ambiguous area of features that for reasons such as specularity, shadows, motion blur, etc., are not located as accurately as other features without clearly being outliers~\citep{triggs2000bundle}.} 
Moreover, outlier rejection methods work best when the majority of data correspondences have small errors, yet most catastrophic failures happen in challenging situations where the majority of data correspondences are unreliable.
Furthermore, while the noise distribution $\errDistSLAM$ is usually modeled to be static, there exist obvious reasons why this is not a realistic assumption. Image features extracted from objects with high-frequency surface textures can be located less accurately; whether the object is static or dynamic affects the observation error; the presence of multiple similar-looking instances of an object in the scene can lead to correspondence errors.

\begin{figure*}[t]
  \centering 
  \includegraphics[width=\linewidth ,trim=0 0 0 0,clip]{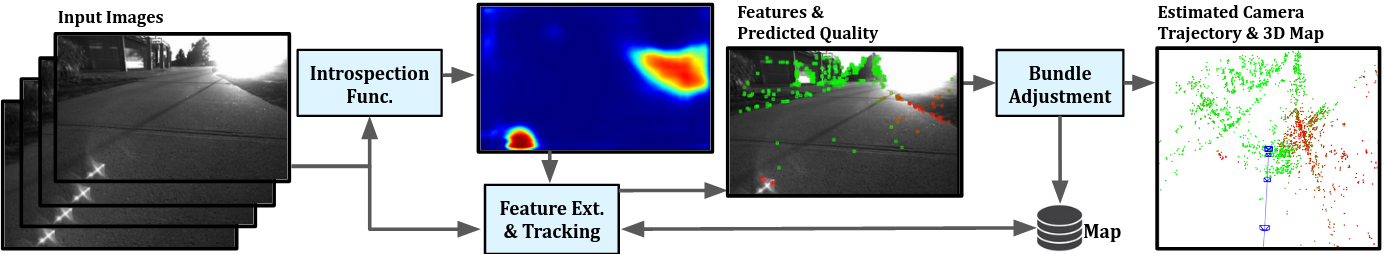}
  \caption{IV-SLAM pipeline during inference.}
  \label{fig:ivslam_pipeline}
  \vspace{-3mm}
\end{figure*}

\subsection{Improving V-SLAM by Learning the Re-projection Error Distribution}
In order to equip V-SLAM with introspection, we apply introspective perception to $\slamFE$ as defined in Eq.~\ref{eq:ipr_def} to learn an approximation of the re-projection error distribution
\begin{align}
\errDistSLAMEst &= \IPrSLAM \quad .
\end{align}
The resultant introspection function estimates the re-projection error distribution, which we use to define the loss function $\mathcal{L}^{\errDistSLAM} \propto -\log\left( \errDistSLAM  \right)$ used by $\slamBE$ to solve for the pose of the camera in Eq.~\ref{eq:bundle_adjustment}.
We select $\mathcal{L}^{\errDistSLAM} \in \mathcal{H}$ where $\mathcal{H}$ is the space of Huber loss functions and specifically
\begin{align}
 \mathcal{L}^{\errDistSLAM}_{\theta( \stateOut{\slamFE}{t,k})}(x) =
 \begin{cases}
             x & \text{if $ x < \theta( \stateOut{\slamFE}{t,k}) $} \\
             2\theta( \stateOut{\slamFE}{t,k})\left(\sqrt{x} - \theta( \stateOut{\slamFE}{t,k})/2 \right) & \text{otherwise}
 \end{cases} 
\label{eq:huber_loss}
\end{align}
where $x = \left\lVert\delta \stateOut{\slamFE}{t, k}\right\rVert^2 \in [0, \infty)$ is the squared $L^2$ norm of the error and $\theta( \stateOut{\slamFE}{t,k})$ is the parameter of the loss function and depends on the reliability of the image feature $\stateOut{\slamFE}{t,k}$.
Introspective V-SLAM (IV-SLAM) uses the introspection function to approximate $\theta( \stateOut{\slamFE}{t,k})$ such that the corresponding error distribution $\errDistSLAMEst$ better models the observed error values. During the training phase, input images and estimated re-projection error values are used to learn to predict the reliability of image features at any point on an image. During the inference phase, a context-aware $\theta( \stateOut{\slamFE}{t,k})$ is estimated for each observed image feature using the predicted reliability score, where a smaller value of $\theta( \stateOut{\slamFE}{t,k})$ corresponds to an unreliable image feature. The resultant loss function $\mathcal{L}^{\errDistSLAM}_{\theta( \stateOut{\slamFE}{t,k})}$ is then used in Eq.~\ref{eq:bundle_adjustment} to solve for $\camPoseAll$ and $\mathbf{M}$. Figure~\ref{fig:ivslam_pipeline} illustrates the IV-SLAM pipeline during inference.

\subsection{Autonomously Supervised Training of IV-SLAM}

\paragraph{Introspection Function Architecture}

We implement the IV-SLAM introspection function as $\IPrSLAM = \IPrSLAMDecompB \circ \IPrSLAMDecompA(\obsPerc{t}{\slamFE})$, where $\IPrSLAMDecompA:\mathbf{I} \to \mathbf{I} $ is a function that given an input image $\obsPerc{t}{\slamFE}$, outputs an image $\costmap$ where $\costmap(i, j) \in [0, 1]$ is the prediction of the normalized reprojection error magnitude for image features extracted at $\obsPerc{t}{\slamFE}(i, j)$. Also $\IPrSLAMDecompB: \mathbf{I} \to \mathbf{I}$  outputs $\lossParamImg$, where $\lossParamImg(i, j)$ is the loss function parameter $\theta$ in Eq.~\ref{eq:huber_loss} for an image feature $\visualFeat{t}{k}$ that is extracted at $\obsPerc{t}{\slamFE}(i, j)$. We implement $\IPrSLAMDecompA(\cdot)$ as a fully convolutional network with the same architecture as that used by~\citet{zhou2017scene} for image segmentation. The network is composed of the MobileNetV2~\citep{sandler2018mobilenetv2} as the encoder and a transposed convolution layer with deep supervision as the decoder.
We train the network using samples of the reprojection error collected in the deployment environment. Moreover, we have $\IPrSLAMDecompB(\costmap(i, j)) = \frac{1 - \costmap(i, j)}{ 1 + \costmap(i, j)} \theta_{\text{max}}$, where $\theta_{\text{max}}$ is a constant positive hyperparameter that defines the range at which $\lossParamImg(i, j)$ varies. In this implementation instead of directly learning to predict $\theta( \stateOut{\slamFE}{t,k})$, IV-SLAM learns to predict the reprojection error magnitude and then computes $\theta( \stateOut{\slamFE}{t,k})$ such that the resultant loss function $\mathcal{L}^{\errDistSLAM}_{\theta( \stateOut{\slamFE}{t,k})}$ more accurately represents the corresponding error distribution. 
For each image feature, the Huber loss is adjusted such that the features that are predicted to be less reliable (larger $\costmap(i, j)$) have a less steep associated loss function (smaller $\theta( \stateOut{\slamFE}{t,k})$).

\begin{figure*}[t]
  \centering 
  \includegraphics[width=\linewidth ,trim=0 0 0 0,clip]{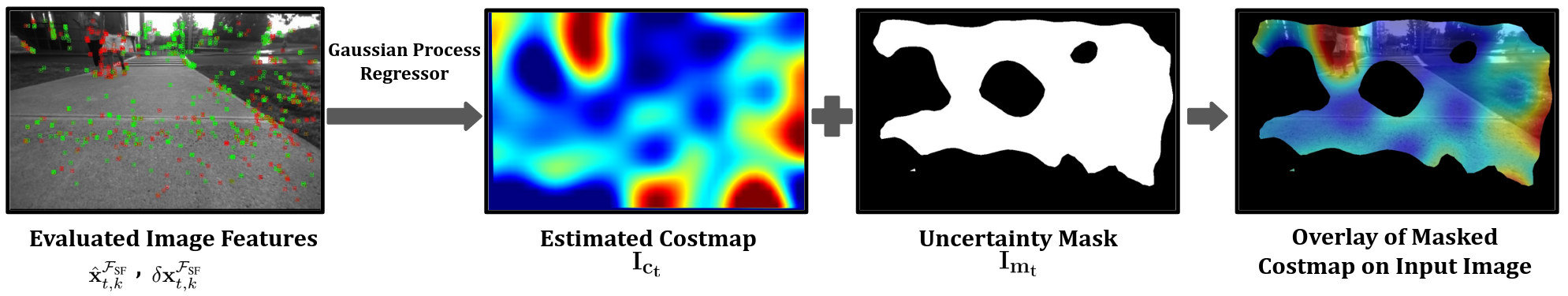}
  \caption{Autonomous training data labeling for the IV-SLAM introspection function.}
  \label{fig:ivslam_training_data}
  \vspace{-3mm}
\end{figure*}

\paragraph{Training Data Collection} \label{sec:ivslam_training}
We use spatio-temporal consistency constraints to collect error samples for $\slamFE$ in the deployment environment and then use the data to train $\IPrSLAMDecompA(\cdot)$ to predict the reprojection error magnitude for every pixel on an input image.
Matched image features are image-space projections of the same object in the world at different time steps. Therefore, they are geometrically related. 
Using the spatio-temporal consistency constraint definition in Eq.~\ref{eq:spatio_temporal_constraint}, we have
\begin{align}
\visualFeat{t}{k} &= \transform{t-\delta t}{t}{\visualFeat{t- \delta t}{k}} \\
 &= \camProject{ \camPoseGen{t - \delta t}{t} \camUnProject{ \visualFeat{t- \delta t}{k}, \featDepth{t-\delta t}{k} }} \, ,
\end{align}
where the transformation function $\transform{t-\delta t}{t}{\cdot}$ is composed of the camera projection function and the relative pose of the camera between the two time steps. In the above equation,
$\camProject{\cdot}: \mathbb{R}^3 \rightarrow \mathbb{P}^2$ is the camera projection function, $\camUnProject{\cdot}: \mathbb{P}^2 \times \mathbb{R} \rightarrow \mathbb{R}^3$ is the inverse, and $\featDepth{t-\delta t}{k}$ is depth of the map point associated with $\visualFeat{t- \delta t}{k}$ in the reference frame of the camera at time $t- \delta t$. 
Following Eq.~\ref{eq:spatio_temporal_consitency}, we use the spatio-temporal consistency constraint to compute the re-projection error for the estimated $\visualFeat{t}{k}$  as
\begin{align}
\reprojErr{t}{k} = \visualFeatEst{t}{k} - \camProject{ \camPoseEstGen{t - \delta t}{t} \camUnProject{ \visualFeatEstPosterior{t- \delta t}{k}{t}, \featDepthEstPosterior{t-\delta t}{k}{t} }} \, ,
\label{eq:ivslam_data_collection}
\end{align}
where $\visualFeatEstPosterior{t- \delta t}{k}{t}$ and $\featDepthEstPosterior{t-\delta t}{k}{t}$ are the posterior estimates of $\visualFeat{t- \delta t}{k}$ and $\featDepth{t-\delta t}{k}$ at time $t$, respectively. 
$ \camUnProject{ \visualFeatEstPosterior{t- \delta t}{k}{t}, \featDepthEstPosterior{t-\delta t}{k}{t}} = \mathbf{p}_k^{t- \delta t} $ is the corresponding map point location in the reference frame of the camera at time $t- \delta t$, and $\camPoseEstGen{t - \delta t}{t}$ is the estimated transformation from the camera frame at time $t - \delta t$ to the camera frame at time $t$.
It should be noted that $\camPoseEstGen{t - \delta t}{t}$ is obtained from solving the bundle adjustment problem (Eq.~\ref{eq:bundle_adjustment}) by the SLAM back-end $\slamBE$ and to ensure that $\camPoseEstGen{t - \delta t}{t}$ is a good approximation for $\camPoseGen{t - \delta t}{t}$, we verify that it agrees with a 3D lidar-based SLAM solution~\citep{shan2018lego} by performing a $\chi^2$ test with $\alpha=0.05$ on the Mahalanobis distance between the two.
Using Eq.~\ref{eq:ivslam_data_collection}, we collect reprojection error samples $\bigl\{ \bigl< \bigl\lVert \reprojErr{t}{k} \bigr\lVert^2 , \visualFeatEst{t}{k} \bigr> \bigr\}_{k=1:N}$ across different images.
In order to prepare the training data for the $\IPrSLAMDecompA(\cdot)$ component of the introspection function, we post-process the set of computed sparse reprojection error values for every image to interpolate the estimated reprojection error for every pixel on the image using a Gaussian Process (GP) regressor.
The outputs of the GP regressor are represented as images $\mathbf{I_c}_t$, which we refer to as cost-maps. 
Then, the pairs of input images and the corresponding cost-maps $\bigl\{ \bigl< \img{t} , \mathbf{I_c}_t \bigr> \bigr\}_{t=1:K}$
are used to train the DNN implementation of $\IPrSLAMDecompA(\cdot)$ using a stochastic gradient descent (SGD) optimizer and a mean squared error loss (MSE) that is only applied to the regions of the image where the uncertainty of the output of the GP is lower than a threshold.
Figure~\ref{fig:ivslam_training_data} shows the estimated cost-map and the uncertainty mask for an example input image. The color of each image feature visualized in the figure corresponds to the magnitude of the actual re-projection error of the map point corresponding to that feature. Green features have a small re-projection error and red features have a large re-projection error. The re-projection error of individual features extracted on the same surface can be different due to the stochastic nature of the noise process in the V-SLAM front-end, the different track lengths of the extracted features, the different scales of the extracted features, and the different 3D locations of the corresponding map points.
In order to train the introspection function, all of these samples of re-projection errors are used and GP regression is used to obtain a smooth estimate of the magnitude of the re-projection error across the image.

\section{Introspective Stereo Depth Estimation}\label{sec:ivoa}

In this section, we present an implementation of introspective perception for a different perception task, stereo depth estimation. We demonstrate how supervisory sensing can be used to learn an introspection function for a depth estimator which in turn is used to predict different types of depth estimate errors.

\subsection{Learning Distribution of Error in Depth Estimates}\label{sec:ivoa-learning}
A stereo depth estimator $\depthEstimator$ outputs a depth image $\depthImgEst{t}$ given a stereo pair of captured images as input
\begin{align}
\depthImgEst{t} = \depthEstimator \bigl( \obsPerc{t}{\depthEstimator} \bigr) \,
\end{align}
where $\obsPerc{t}{\depthEstimator} = \left< \leftImg{t}, \rightImg{t} \right>$ and $\leftImg{t}$ and $\rightImg{t}$ are the images captured by the left and right cameras, respectively. Moreover, $\depthImgEst{t}(u, v)$ is the estimated depth for the pixel coordinates $(u, v)$ in the left image. Depth estimate error for each pixel location is defined as 
\begin{align}
\depthEstimateError{t} = \depthImgEst{t}(u, v) - \depthImg{t}(u, v) \, ,
\end{align}
where $\depthEstimateError{t} \sim \errDistDepth\bigl( \obsPerc{t}{\depthEstimator}, \left<u, v\right> \bigr)$. The error distribution $\errDistDepth$ is dependent on the context at every point on the image. The texture, shape, and surface material of the objects in the image combined with different lighting conditions can lead to different error distributions across frames and in different regions in the same image.

Following the definition of introspective perception in Eq.~\ref{eq:ipr_def}, we equip a stereo depth estimator with introspection by learning an approximation of the depth error distribution 
\begin{align}
\errDistDepthEst = \IPrDepth \, ,
\end{align} 
where $\errDistDepthEst$ is a piece-wise approximation of $\errDistDepth$ such that
\begin{align}
 \errDistDepthEst_{\theta{\bigl( \obsPerc{t}{\depthEstimator}   , \left<u, v\right> \bigr)}} (x) =
 \begin{cases}
             \theta_{\text{FP}}{\bigl( \obsPerc{t}{\depthEstimator}   , \left<u, v\right> \bigr)} &    -R_{\max} < x < -\alpha \\
             \theta_{\text{T}}{\bigl( \obsPerc{t}{\depthEstimator}   , \left<u, v\right> \bigr)}  &   -\alpha \leq x \leq \alpha \\
              \theta_{\text{FN}}{\bigl( \obsPerc{t}{\depthEstimator}  , \left<u, v\right> \bigr)} &   \alpha < x < R_{\max}
 \end{cases} 
\label{eq:depthErrDist}
\end{align}
where $\alpha$ is a constant error threshold, $R_{\max}$ is the maximum range of the depth estimator, and $\theta_{\text{FN}}$, $\theta_{\text{T}}$, and $\theta_{\text{FP}}$ are the parameters of the error distribution and are learned. The probability density value $\theta_{\text{FN}}$ corresponds to cases when the estimated depth is larger than the actual depth, which we refer to as instances of false negatives (FN). Moreover, probability density value $\theta_{\text{FP}}$ corresponds to cases when the estimated depth is smaller than the actual depth, which we refer to as instances of false positives (FP).
Using such piece-wise approximation of $\errDistDepth$ rather than a continuous probability density function has the benefit of being easier to learn, while still being informative for predicting failures at the task level.

\subsection{Autonomously Supervised Training of the Introspection Function}

\paragraph{Introspection Function Architecture} 
We implement $\IPrShortest(\depthEstimator)$ as a convolutional neural network (CNN) with the same architecture as AlexNet~\citep{krizhevsky2012imagenet}. The input to the network are image patches $\leftImg{t, \left<u, v\right>}$ extracted from different pixel locations $\left<u, v\right>$ in the left camera images. The output are predicted probability scores $
\left<p_{\text{FP}}, p_{\text{FN}}, p_{\text{TP}}, p_{\text{TN}} \right>$ for each of the four different classes of false positive (FP), false negative (FN), true positive (TP), and true negative (TN) which make up the parameters of $\errDistDepthEst_{\theta{\bigl( \obsPerc{t}{\depthEstimator}   , \left<u, v\right> \bigr)}}$ such that $\theta_{\text{FP}} = \frac{p_{\text{FP}}}{R_{\max} - \alpha}$ , $\theta_{\text{FN}} = \frac{p_{\text{FN}}}{R_{\max} - \alpha}$, and $\theta_{\text{T}} = \frac{p_{\text{TN}} + p_{\text{TP}}}{2\alpha}$. 

\paragraph{Training Data Collection}
We use consistency across sensors to collect samples of depth estimation error values. Specifically, we use a Kinect depth camera as a supervisory sensor and use its depth estimations $  \depthImgEstSupervisory{t} = \depthEstimatorSupervisory \bigl( \obsPerc{t}{\depthEstimatorSupervisory} \bigr)$ to obtain the reference depth values for the stereo depth estimator $\depthEstimator$

\begin{align}
\depthImg{t} = \camProjectForFrame{ \camUnProjectForFrame{\depthImgEstSupervisory{t}}{\text{D}^{\prime}} }{\text{D}} \, ,
\end{align}
where $\camUnProjectForFrame{\cdot}{\text{D}^{\prime}}: \mathbb{P}^2 \times \mathbb{R} \rightarrow  \mathbb{R}^3$ unprojects pixels from the Kinect depth image into the 3D robot local frame  and $\camProjectForFrame{\cdot}{\text{D}}: \mathbb{R}^3 \rightarrow \mathbb{P}^2 \times \mathbb{R}$ projects 3D points from the robot frame onto the primary camera images and assigns a depth value to each pixel.
Following Eq.~\ref{eq:sensing_consitency}, we use consistency across sensors to compute the depth estimation errors as
\begin{align}
\depthImgError{t} &= \depthImgEst{t} - \camProjectForFrame{ \camUnProjectForFrame{\depthImgEstSupervisory{t}}{\text{D}^{\prime}} }{\text{D}} \, ,
\end{align}
where $\depthEstimateError{t} = \depthImgError{t}(u, v)$. The robot collects samples of depth estimate errors during deployment when the supervisory sensor is available, e.g. indoors. This data is then used to train the introspection function.

\section{Experimental Results}\label{sec:results}
In this section we evaluate our implementations of introspective perception for both V-SLAM and stereo depth estimation. We 1) evaluate the accuracy of the learned introspection functions in predicting errors of perception, 2) assess the effect of introspective perception in reducing failures of autonomy, 3) investigate the use of introspective perception in identifying the causes of robot failures, and 4) evaluate the performance of introspective perception, when the underlying perception algorithm is fully-learned and compare it against SOTA methods for estimating the uncertainty of deep neural networks.

We implement and evaluate \IPrName{} for three different perception algorithms. 1) We implement IV-SLAM for the stereo version of ORB-SLAM~\citep{mur2017orb}, a popular optimization-based visual SLAM algorithm. 2) We implement introspective depth estimation for JPP-C~\citep{ghosh2017joint}, which is a fast and computationally efficient geometric method for sparse stereo reconstruction. 3) We also implement introspective depth estimation for GA-Net~\citep{zhang2019ga}, which is a deep-learning-based approach for stereo depth estimation.

\subsection{Evaluation Datasets}

\subsubsection{V-SLAM Dataset} \label{ivslam_dataset}
State-of-the-art vision-based SLAM algorithms have shown great performance on popular benchmark datasets such as KITTI~\citep{geiger2012we} and EuRoC~\citep{burri2016euroc}. These datasets, however, do not reflect the many difficult situations that can happen when the robots are deployed in the wild and over extended periods of time~\citep{shi2019we}. In order to assess the effectiveness of IV-SLAM on improving visual SLAM performance, in addition to evaluation on the public EuRoC and KITTI datasets, we also put IV-SLAM to test on simulated and real-world datasets that we have collected to expose the algorithm to challenging situations such as reflections, glare, shadows, and dynamic objects.

\paragraph{Simulation}
In order to evaluate IV-SLAM in a controlled setting, where we have accurate poses of the camera and ground truth depth of the scene, we use AirSim~\citep{shah2018airsim}, a photo-realistic simulation environment. A car is equipped with a stereo pair of RGB cameras as well as a depth camera that provides ground truth depth readings for every pixel in the left camera frame. The dataset encompasses more than \SI{60}{\kilo\meter} traversed by the car in the City environment under different weather conditions such as clear weather, wet roads, and in the presence of snow and leaves accumulation on the road.

\paragraph{Real-world} 
We also evaluate IV-SLAM on real-world data that we collect using a Clearpath Jackal robot equipped with a stereo pair of RGB cameras as well as a Velodyne VLP-16 3D Lidar. The dataset consists of more than \SI{7}{\kilo\meter} worth of trajectories traversed by the robot over the span of a week in a college campus setting in both indoor and outdoor environments and different lighting conditions.
The reference pose of the robot is estimated by LeGO-LOAM~\citep{shan2018lego}, a 3D Lidar-based SLAM solution. The algorithm is run on the data offline and with extended optimization rounds for increased accuracy.
For both the real-world and simulation datasets the data is split into training and test sets, each composed of separate full deployment sessions(uninterrupted trajectories), such that both training and test sets include data from all environmental conditions.

\subsubsection{Stereo Depth Estimation Dataset}
We implement and evaluate introspective perception for two different stereo depth estimation algorithms: JPP-C~\citep{ghosh2017joint}, which is a model-based approach, and GA-Net~\citep{zhang2019ga}, which is an end-to-end learning approach. For the former, we evaluate the approach on a real-robot dataset. For the latter, we evaluate the approach in simulation, where we can easily collect sufficient data for training the underlying fully-learned perception algorithm. 

\paragraph{Simulation}\label{sec:depth_dataset_sim}
We use the AirSim simulator with the same sensor configuration as that used for the V-SLAM dataset. For the stereo depth estimation dataset, in addition to the City environment, we also collect data in the Neighborhood environment. This additional environment is held out during the training of the underlying fully-learned perception algorithm and used only for testing to evaluate the performance of \IPrName{} on out of distribution (OOD) data.

\paragraph{Real-world}
The real-world dataset is collected on a Clearpath Jackal with a similar sensor suite to that used for collecting the V-SLAM dataset. The robot is deployed in both indoor and outdoor settings and the RGB images captured by the stereo pair of Point Grey cameras as well as depth images captured by a Kinect depth sensor are recorded at \SI{30}{\Hz}.
The data is processed offline and the Kinect depth images are registered to the RGB images from the left camera to provide reference depth estimates. The indoor dataset spans multiple buildings with different types of tiling and carpet. The outdoor dataset is also collected on different surfaces such as asphalt, concrete, and tile in both dry and wet conditions. The total dataset that is generated from more than \SI{1.4}{\kilo\meter} robot deployments
includes about $120$k image frames. The data is split into train and test datasets, each composed of separate full robot deployment sessions, such that both train and test sets include data from all types of terrain.

\subsection{Introspection Function Accuracy}

\begin{figure}
  \centering
  \includegraphics[width=0.45\linewidth]{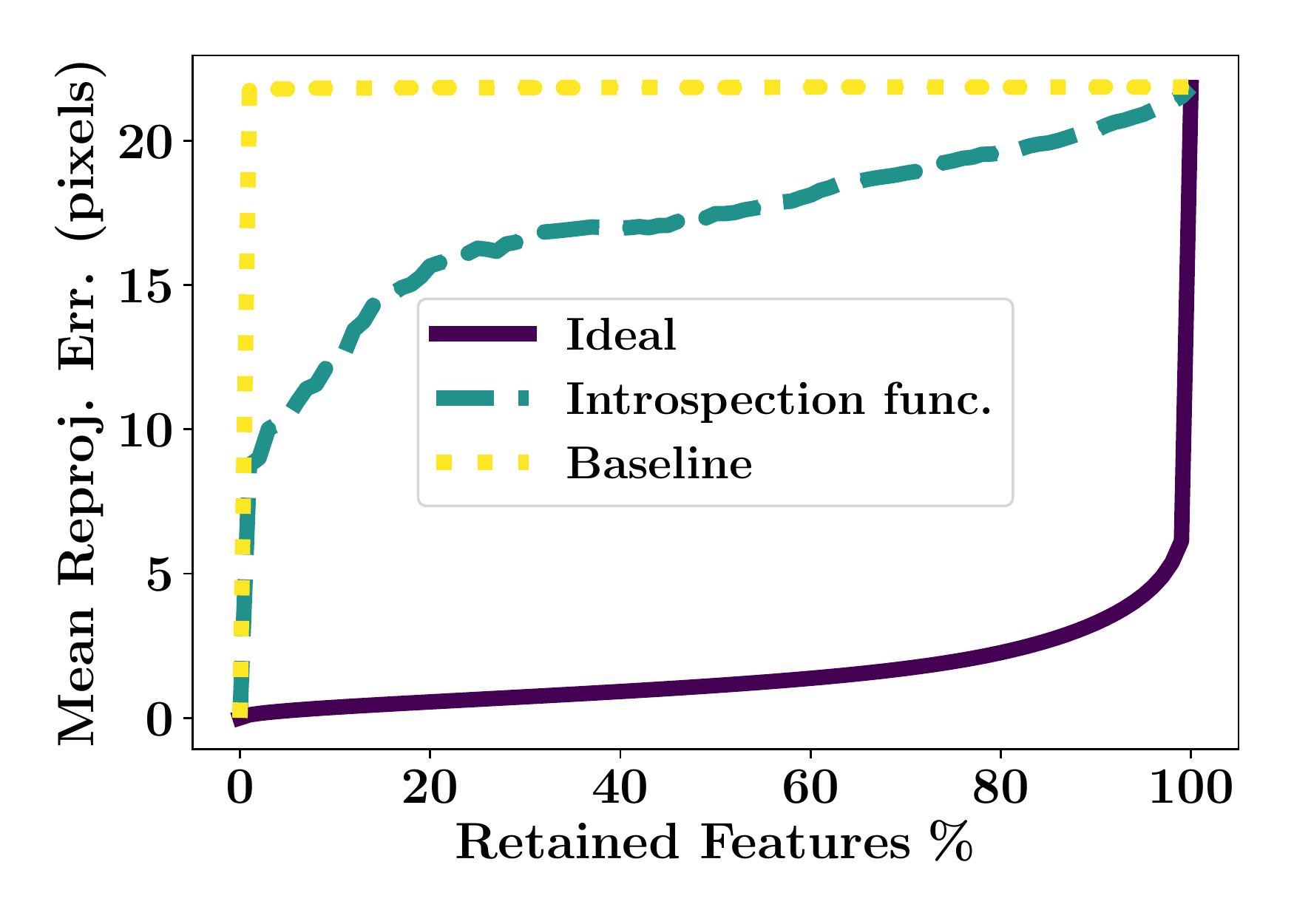}
  \caption{Mean reprojection error for image features sorted
  1) randomly (baseline),
  2) based on predicted reliability (Introspection func.),
  3) based on ground truth reprojection error (ideal)
  }\label{fig:feature_qual_pred}
\end{figure}

\begin{figure}[t]
  \begin{subfigure}[b]{0.194\linewidth}
    \includegraphics[width=1\linewidth, trim=0 0 0 0,clip]{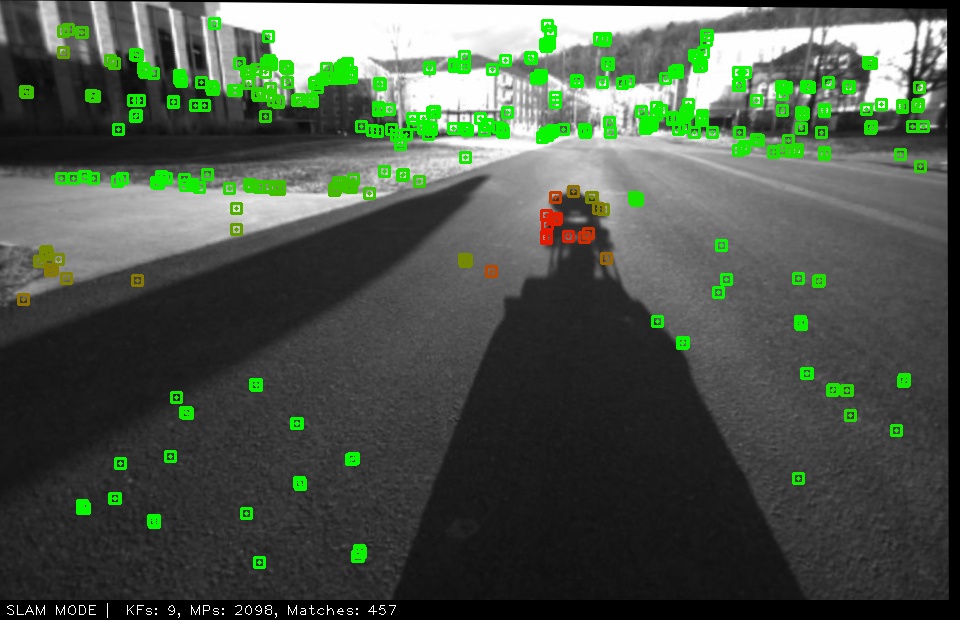}
  \end{subfigure} 
  \begin{subfigure}[b]{0.194\linewidth}
    \includegraphics[width=\linewidth, trim=0 0 0 0,clip]{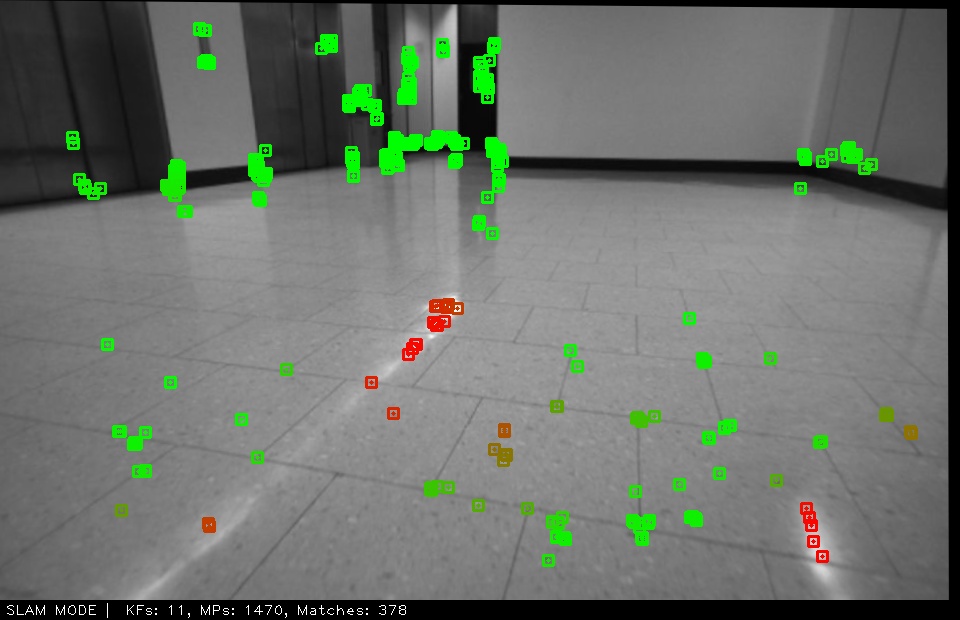}
  \end{subfigure} 
  \begin{subfigure}[b]{0.194\linewidth}
    \includegraphics[width=\linewidth, trim=0 0 0 0,clip]{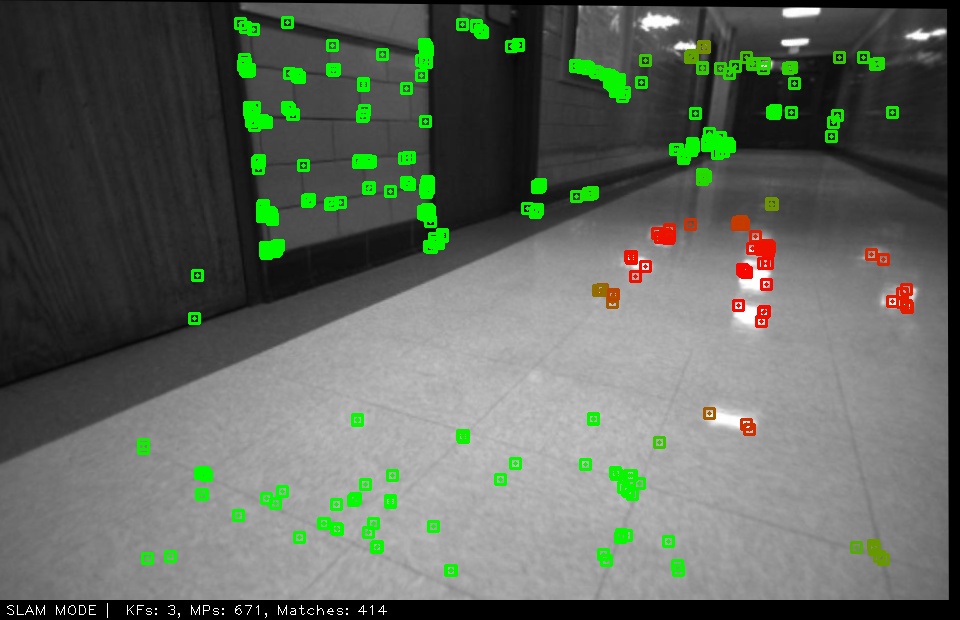}
  \end{subfigure} 
  \begin{subfigure}[b]{0.194\linewidth}
    \includegraphics[width=\linewidth, trim=0 0 0 0,clip]{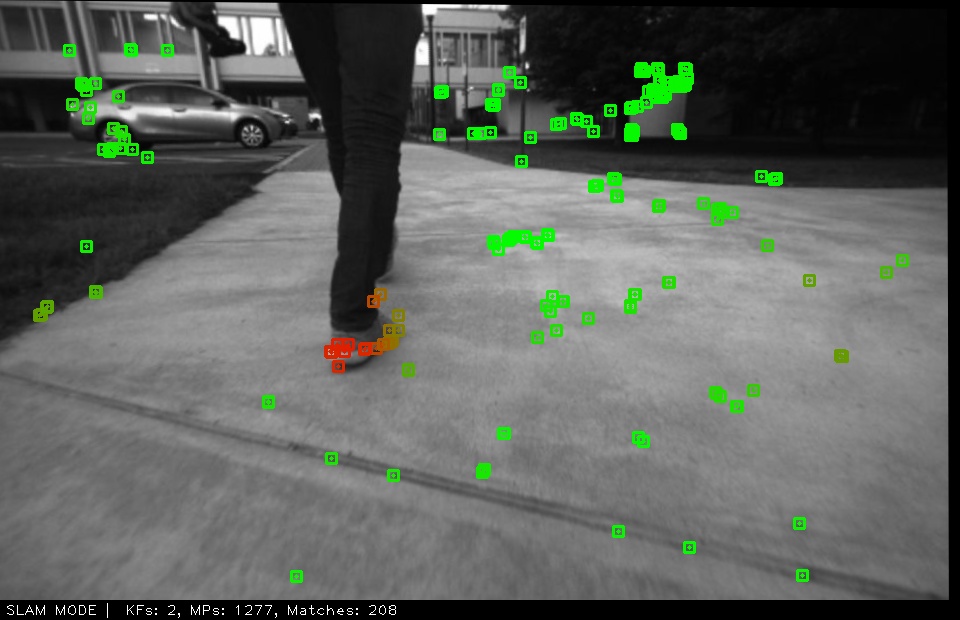}
  \end{subfigure}
  \begin{subfigure}[b]{0.194\linewidth}
    \includegraphics[width=\linewidth, trim=0 0 0 0,clip]{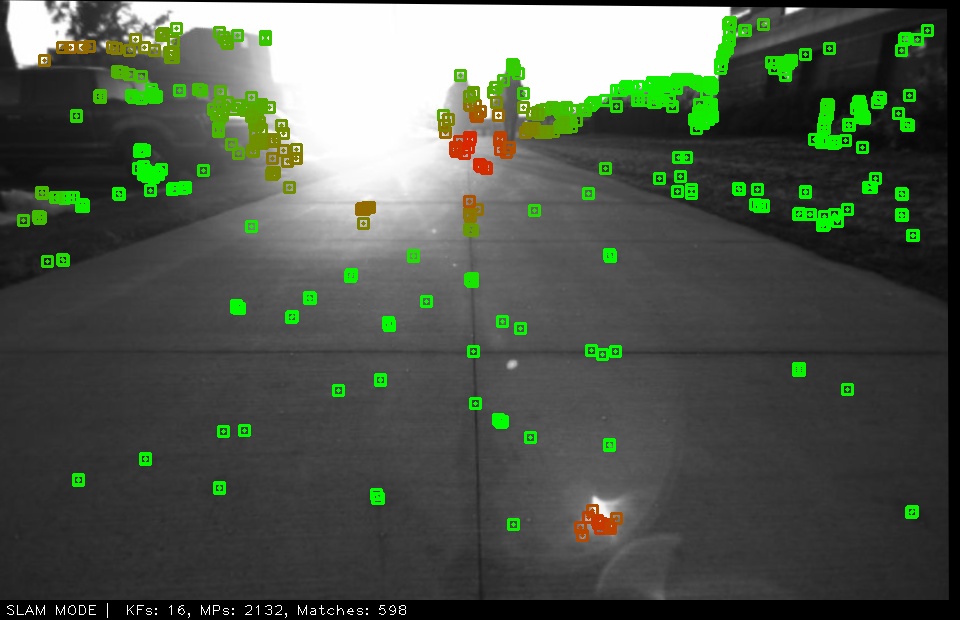}
  \end{subfigure}
   \begin{subfigure}[b]{0.194\linewidth}
    \includegraphics[width=\linewidth, trim=0 0 0 0,clip]{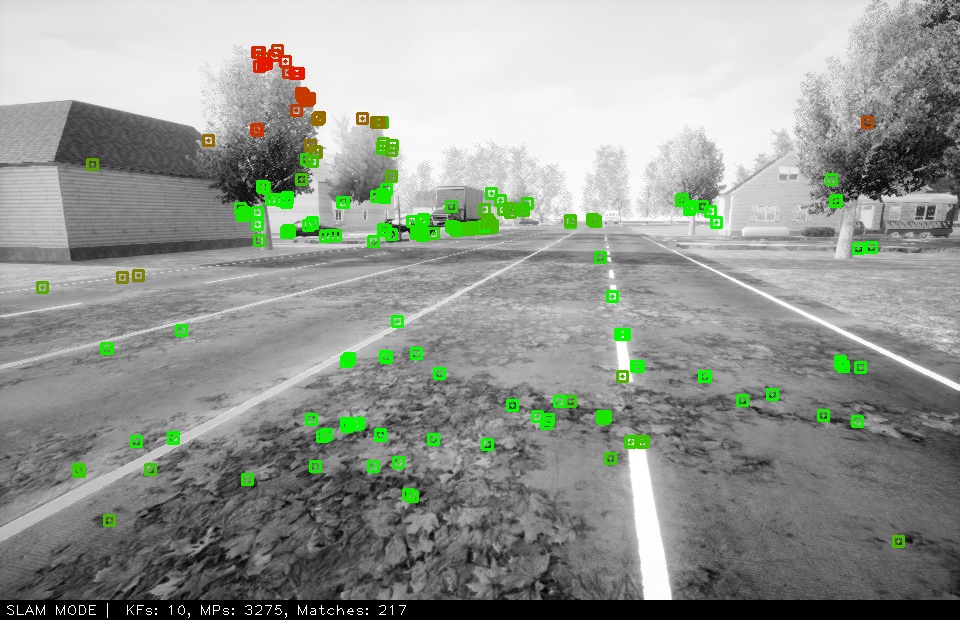}
  \end{subfigure} 
  \begin{subfigure}[b]{0.194\linewidth}
    \includegraphics[width=\linewidth, trim=0 0 0 0,clip]{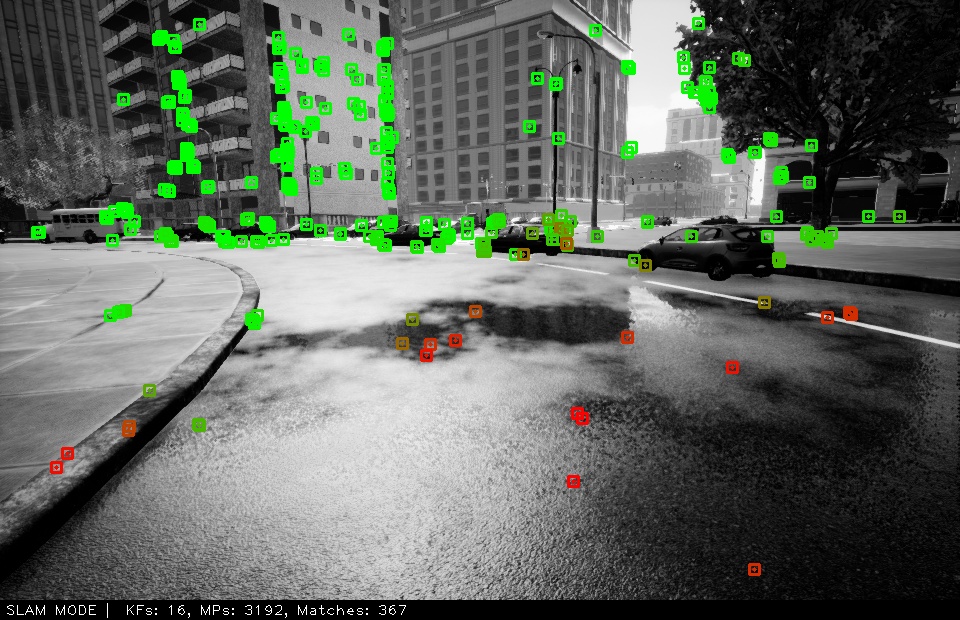}
  \end{subfigure} 
  \begin{subfigure}[b]{0.194\linewidth}
    \includegraphics[width=\linewidth, trim=0 0 0 0,clip]{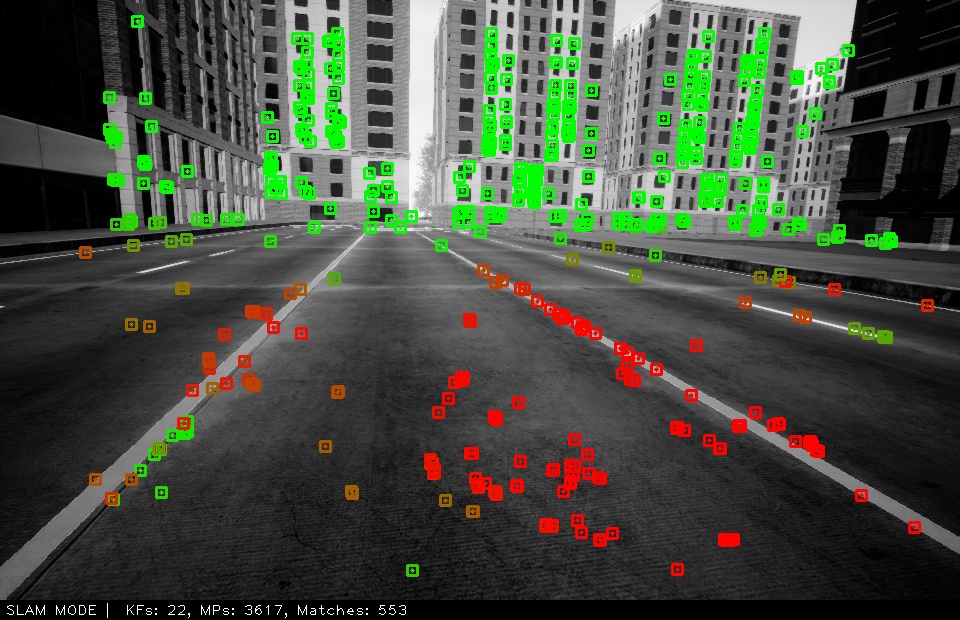}
  \end{subfigure} 
  \begin{subfigure}[b]{0.194\linewidth}
    \includegraphics[width=\linewidth, trim=0 0 0 0,clip]{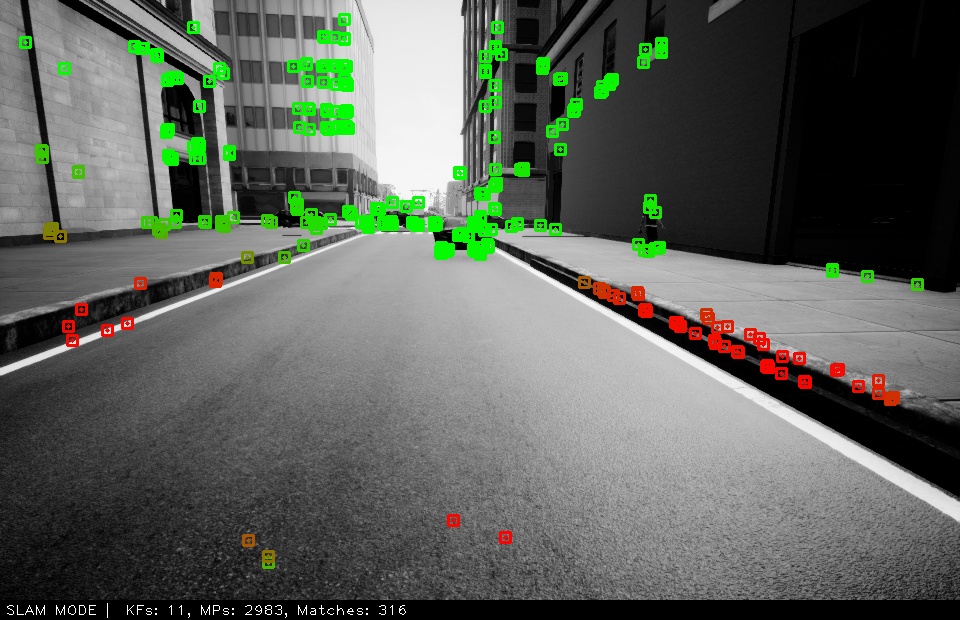}
  \end{subfigure}
  \begin{subfigure}[b]{0.194\linewidth}
    \includegraphics[width=\linewidth, trim=0 0 0 0,clip]{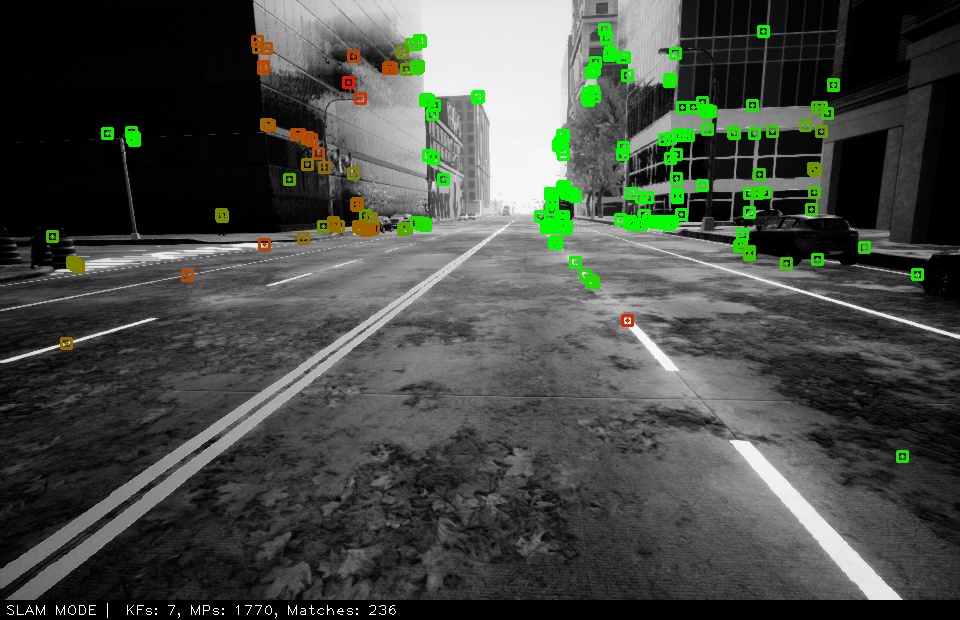}
  \end{subfigure}
 
  \caption{Snapshots of IV-SLAM running on real-world data (top row) and in simulation (bottom row). Green and red points on the image represent the reliable and unreliable tracked image features, respectively, as predicted by the introspection function. Detected sources of error include shadow of the robot, surface reflections, pedestrians, glare, and ambiguous image features extracted from high-frequency textures such as asphalt or vegetation.}
  \label{fig:image_snapshots}
 \end{figure}

\paragraph{IV-SLAM Introspection Function}
We evaluate the IV-SLAM introspection function based on its performance to correctly predict reliability of image features. We expect image features $\visualFeat{t}{k}$ for which a smaller loss function parameter $\theta(\visualFeat{t}{k})$ is predicted, to have larger reprojection errors (lower reliability).
Since obtaining the ground truth reprojection error requires access to ground truth 3D coordinates of objects associated with each image feature as well as accurate reference poses of the camera, we conduct this experiment in simulation.
IV-SLAM is trained on the simulation dataset with the method explained in \cref{sec:ivslam_training}. The introspection function is then run on all images in the test set along with the original ORB-SLAM. For each image feature extracted and matched by ORB-SLAM, we log its corresponding ground truth reprojection error, as well as the predicted loss function parameter by the introspection function.
We then sort all image features in ascending order with respect to \begin{inparaenum}[1)]
\item ground truth reprojection errors and
\item the negative of the predicted loss function parameter $\theta(\visualFeat{t}{k})$.
\end{inparaenum}
Figure~\ref{fig:feature_qual_pred} illustrates the mean reprojection error for the top $x\%$ of features in each of these ordered lists for a variable $x$. The lower the area under a curve in this figure, the more accurate is the corresponding image feature evaluator in sorting features based on their reliability. The curve corresponding to the ground truth reprojection errors indicates the ideal scenario where all image features are sorted correctly. The baseline is also illustrated in the figure as the resultant mean reprojection error when the features are sorted randomly (mean error over 1000 trials) and it corresponds to the case when no introspection is available and all features are treated equally. As can be seen, using the introspection function significantly improves image feature reliability assessment.
Figure~\ref{fig:image_snapshots} illustrates the output of the introspection function for image features extracted at different scenes. The function attributes higher uncertainty to features extracted from regions that contain shadow of the robot, surface reflections, lens flare, and pedestrians.
It is noteworthy that IV-SLAM has learned to identify these different sources of V-SLAM error autonomously and without any pre-enumeration of the different types of errors. This highlights the advantage of using introspection as opposed to hand-crafted error detection methods. Hand-crafted and heuristic metrics for evaluating the reliability of image features target a specific subset of the sources of image-matching errors by design and do not adapt to predict new sources of image-matching errors. For instance, while spatial entropy can be used to detect instances of textureless or blurry images~\citep{brunner2013selective}, this metric does not predict the higher reprojection error on the crisp corners of the shadow of a moving object, for which we have observed V-SLAM failures. However, we have shown that such sources of error that are correlated with the contextual and semantic information in the input image can be discovered by introspective perception, and the autonomously supervised training of the introspective perception enables adapting to previously unseen sources of errors, removing the need for pre-enumeration of the error sources.

 \begin{figure}[t]
  \centering
   \begin{subfigure}[b]{0.41\linewidth}
    \includegraphics[width=1.0\linewidth,trim=0 0 0 0,clip]{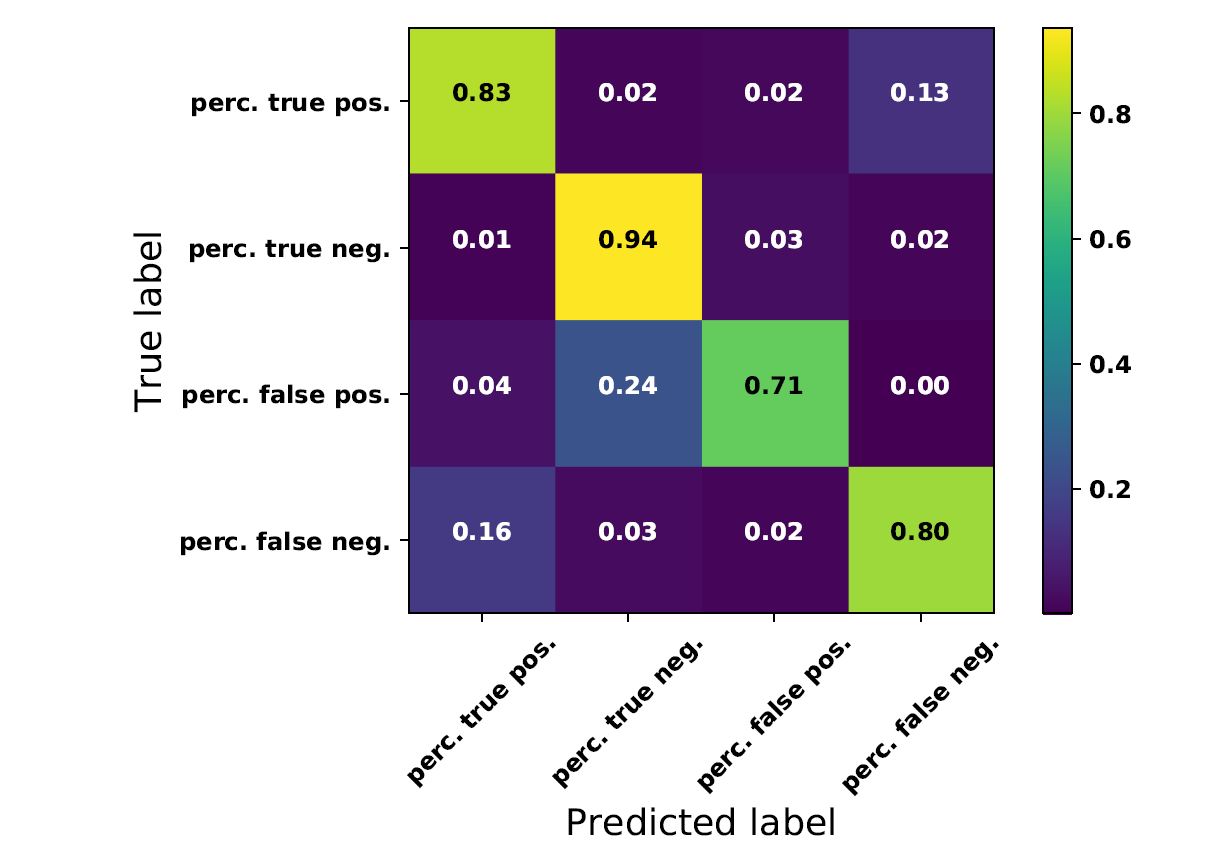}
    \caption{}
    \label{fig:agg_results_indoor}
   \end{subfigure}
   \begin{subfigure}[b]{0.41\linewidth}
    \includegraphics[width=1.0\linewidth,trim=0 0 0 0,clip]{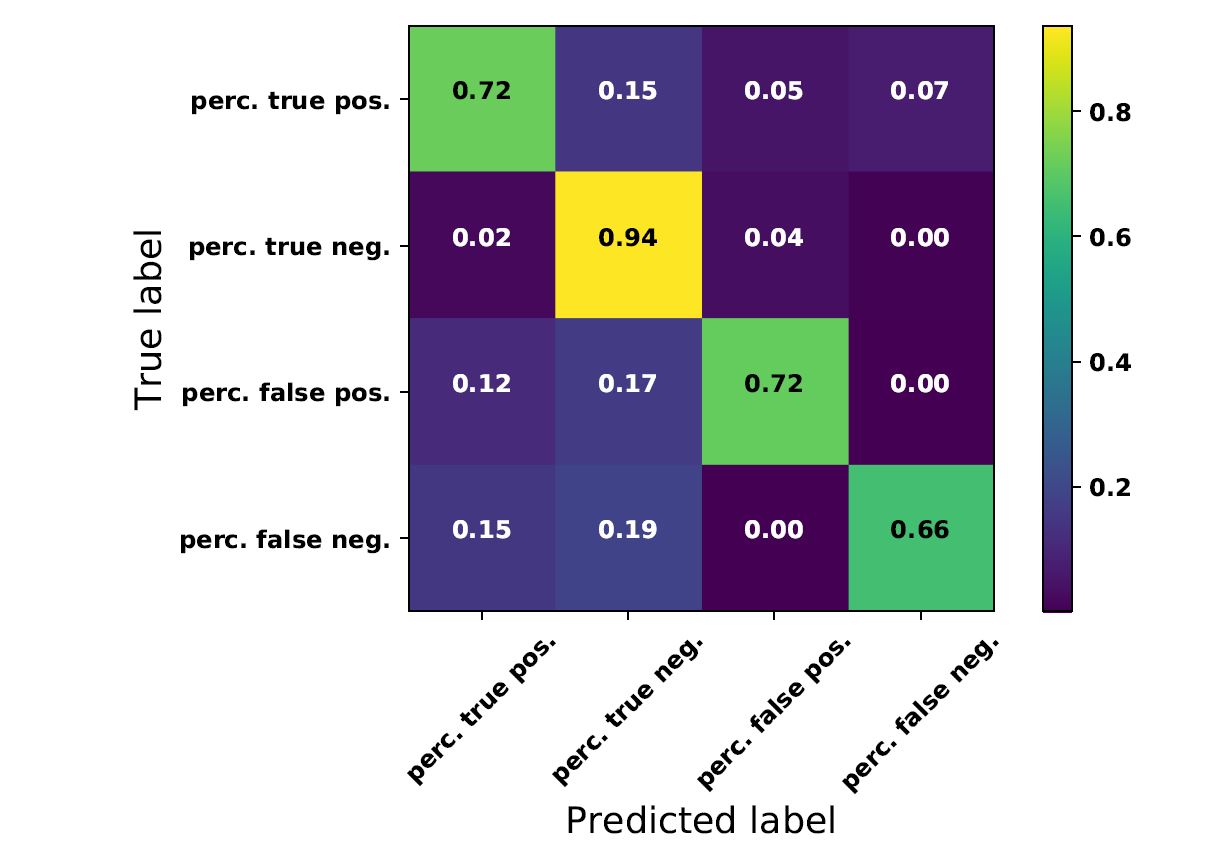}
    \caption{}
    \label{fig:agg_results_outdoor}
   \end{subfigure}
    \caption{Prediction results of the depth estimator introspection function on the (\subref{fig:agg_results_indoor}) indoor and (\subref{fig:agg_results_outdoor}) outdoor datasets.}
   \label{fig:classification_report}
  \end{figure}

  \begin{figure}[t]
    \centering 
    \includegraphics[width=0.5\linewidth,trim=0 0 0 0,clip]{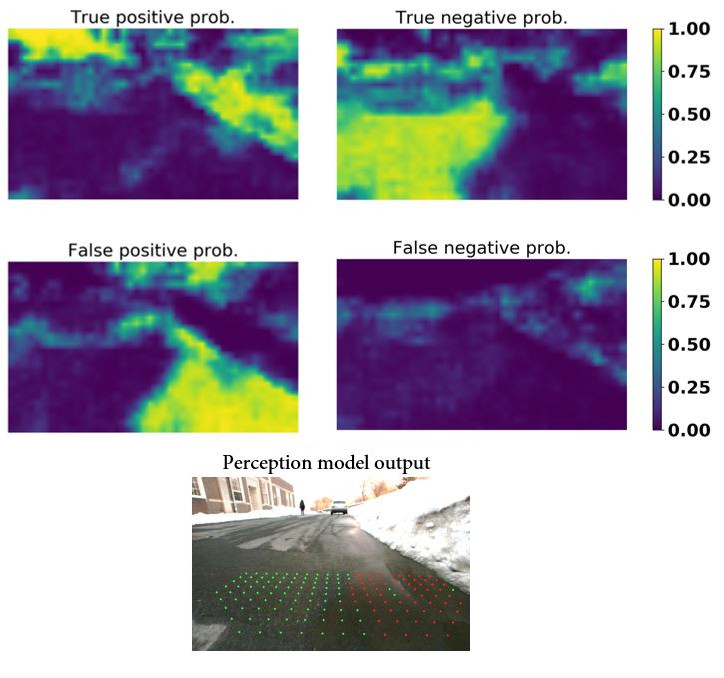}
    \caption{Example output of the depth estimator introspection function. The bottom image shows the estimated obstacle grid by the depth estimator
  in front of the robot, where the red and green dots denote the detected occupied and obstacle free cells, respectively. The introspection function correctly predicts water reflections to cause false positives for the perception model (middle left image).}
    \label{fig:detailed_introspection_output}
  \end{figure}

\paragraph{Depth Estimator Introspection Function}
We evaluate the depth estimator introspection function based on its performance in predicting the occurrence of depth estimation errors and their types. Figure~\ref{fig:classification_report} shows the prediction results. The introspection function $\IFShort{\depthEstimator}$ can predict the different types of errors with reasonably high accuracy in both indoor and outdoor settings. Figure~\ref{fig:detailed_introspection_output} also illustrates the output of $\IFShort{\depthEstimator}$ for an example input image, and shows how the introspection function correctly predicts water puddle reflections to lead to depth estimate errors.

\subsection{Effect of Introspective Perception on Robustness in Autonomy}

As explained in \cref{sec:ivslam}, IV-SLAM improves the robustness of V-SLAM by using the output of the introspection function, when solving for the pose of the camera in the SLAM back-end.
We compare our introspection-enabled version of ORB-SLAM with the original algorithm in terms of their camera pose tracking accuracy and robustness. Both algorithms are run on the test data and their estimated poses of the camera are recorded.
If the algorithms lose track due to lack of sufficient feature matches across frames, tracking is reinitialized and continued from after the point of failure along the trajectory and the event is logged as an instance of tracking failure for the corresponding SLAM algorithm.
The relative pose error (RPE) is then calculated for both algorithms at consecutive pairs of frames that are $d$ meters apart as defined in~\citep{schubert2018tum}. 

\begin{table}[]
  \parbox{.45\linewidth}{
  \caption{\textsc{Tracking Accuracy in the KITTI Dataset}}
  \label{table:kitti_results}
  \centering
  \resizebox{0.5\textwidth}{!}{
  \begin{tabular}{l c c c c }
    \toprule
    \multicolumn{1}{l}{} &
  	\multicolumn{2}{c}{IV-SLAM} &
  	\multicolumn{2}{c}{ORB-SLAM} \\
    \cmidrule(r{0.7em}){2-3}
	\cmidrule(l{0.0em}){4-5}    
    
  	\multirow[b]{2}{1.25cm}{Sequence} &
 	\multirow[b]{2}{1.5cm}{\centering Trans. Err. \%} &
 	\multirow[b]{2}{2.0cm}{\centering Rot. Err. (\si{deg/100m})} &
 	\multirow[b]{2}{1.5cm}{\centering Trans. Err. \%} &
 	\multirow[b]{2}{2.0cm}{\centering Rot. Err. (\si{deg/100m})} \\
 	& & & &  \\
    \midrule
 00  & 0.69 		   & 0.25		     & 0.69	                & 0.25  \\
 01  & \textbf{1.43}   & 0.22 		     & 1.47 		        & 0.22  \\
 02  & 0.79		       & \textbf{0.22} 	 & \textbf{0.76}	    & 0.24  \\
 03  & 0.74		       & \textbf{0.19}   & \textbf{0.70} 	    & 0.23  \\
 04  & \textbf{0.49}   & 0.13 		     & 0.55			        & 0.13 \\
 05  & 0.40 		   & 0.16	         & \textbf{0.38}	    & 0.16  \\
 06  & \textbf{0.49}   & \textbf{0.14}   & 0.56				    & 0.19  \\
 07  & 0.49            & \textbf{0.27}   & 0.49				    & 0.29 \\
 08  & \textbf{1.02}   & 0.31 	         & 1.05  				& 0.31  \\
 09  & 0.85 		   & 0.25	         & \textbf{0.82} 		& 0.25 \\
 10  & \textbf{0.61}   & \textbf{0.26}   & 0.62 				& 0.29 \\
    \midrule
 Average  & 0.77       & \textbf{0.24} & 0.77 		        & 0.25 \\
    \bottomrule
  \end{tabular}
    }
  } \qquad
  \parbox{.45\linewidth}{
    \centering
    \caption{\textsc{Tracking Accuracy in the EuRoC Dataset}}
    \label{table:euroc_results}
    \resizebox{0.5\textwidth}{!}{
    \begin{tabular}{l c c c c }
      \toprule
      \multicolumn{1}{l}{} &
      \multicolumn{2}{c}{IV-SLAM} &
      \multicolumn{2}{c}{ORB-SLAM} \\
      \cmidrule(r{0.7em}){2-3}
    \cmidrule(l{0.0em}){4-5}    
      
      \multirow[b]{2}{1.25cm}{Sequence} &
     \multirow[b]{2}{1.5cm}{\centering Trans. Err. \%} &
     \multirow[b]{2}{2.0cm}{\centering Rot. Err. (\si{deg/m})} &
     \multirow[b]{2}{1.5cm}{\centering Trans. Err. \%} &
     \multirow[b]{2}{2.0cm}{\centering Rot. Err. (\si{deg/m})} \\
     & & & &  \\
      \midrule
   MH1  & \textbf{2.26} 		   & \textbf{0.19}		              & 2.42	    & 0.21  \\
   MH2  & 1.78                   & 0.18 		     & \textbf{1.49} 		    & \textbf{0.16} \\
   MH3  & 3.27		           & 0.18 	         & 3.27	                   & \textbf{0.17}  \\
   MH4  & 3.85		           & 0.16            & \textbf{3.49} 	          & \textbf{0.15}  \\
   MH5  & \textbf{2.98}          & \textbf{0.16} 		     & 3.32			               & 0.18 \\
   V1\_1  & 8.93 		           & \textbf{1.05}	         & \textbf{8.85}	             & 1.06  \\
   V1\_2  & \textbf{4.38}         & 0.41            & 4.46				    & \textbf{0.39}  \\
   V1\_3  & \textbf{7.85}         & \textbf{1.24}            & 14.86			    & 2.35 \\
   V2\_1  & \textbf{2.92}         & \textbf{0.76}	         & 4.37  				& 1.39  \\
   V2\_2  & 2.89 		            & 0.62	         & 2.76 		        & \textbf{0.59} \\
   V2\_3  & \textbf{11.00}         & 2.49            & 12.73 				& \textbf{2.39} \\
      \midrule
   Average  & \textbf{4.74}       & \textbf{0.68}            & 5.64 		        & 0.82 \\
      \bottomrule
    \end{tabular}
    }
  }
\end{table}

\begin{table}[t]
  \caption{\textsc{Aggregate Results for Simulation and Real-world Experiments}}
  \label{table:aggregate_results}
  \centering
  \resizebox{\textwidth}{!}{%
  \begin{tabular}{l c c c c c c }
    \toprule
    \multicolumn{1}{l}{} &
  	\multicolumn{3}{c}{Real-World} &
  	\multicolumn{3}{c}{Simulation} \\
    \cmidrule(r{0.7em}){2-4}
	\cmidrule(l{0.0em}){5-7}    
    
  	\multirow[b]{1}{2.0cm}{Method} &
 	\multirow[b]{1}{3.0cm}{\centering Trans. Err. \%} &
 	\multirow[b]{1}{3.0cm}{\centering Rot. Err. (\si{deg/m})} &
 	\multirow[b]{1}{2.0cm}{\centering MDBF (\si{m})} &
 	\multirow[b]{1}{3.0cm}{\centering Trans. Err. \%} &
 	\multirow[b]{1}{3.0cm}{\centering Rot. Err. (\si{deg/m})} &
    \multirow[b]{1}{2.0cm}{\centering MDBF (\si{m})} \\
    \midrule
 IV-SLAM  & \textbf{5.85} & \textbf{0.511} & \textbf{621.1} & \textbf{12.25} & \textbf{0.172} & \textbf{450.4} \\
 ORB-SLAM  & 9.20 & 0.555 & 357.1 & 18.20 & 0.197 & 312.7 \\
    \bottomrule
  \end{tabular}}
\end{table}

\paragraph{KITTI and EuRoC datasets}
We perform leave-one-out cross-validation separately on the KITTI and EuRoC datasets, i.e. to test IV-SLAM on each sequence, we train it on all other sequences in the same dataset. 
Tables~\ref{table:kitti_results} and \ref{table:euroc_results} compare the per trajectory root-mean-square error (RMSE) of the rotation and translation parts of the RPE for IV-SLAM and ORB-SLAM in the KITTI and EuRoC datasets, respectively.
The baseline ORB-SLAM does a good job of tracking in the KITTI dataset. There exists no tracking failures and the overall relative translational error is less than $1\%$. Given the lack of challenging scenes in this dataset, IV-SLAM performs similar to ORB-SLAM with only marginal improvement in the overall rotational error. While EuRoC is more challenging than the KITTI given the higher mean angular velocity of the robot, the only tracking failure happens in the V2\_3 sequence and similarly for both ORB-SLAM and IV-SLAM due to severe motion blur. IV-SLAM performs similar to ORB-SLAM on the easier trajectories, however, it significantly reduces the tracking error on the more challenging trajectories such as V1\_3. Over all the sequences, IV-SLAM improves both the translational and rotational tracking accuracy.

\begin{figure}[t]
  \begin{subfigure}[b]{0.32\linewidth}
    \includegraphics[width=1\linewidth, trim=0 0 0 0,clip]{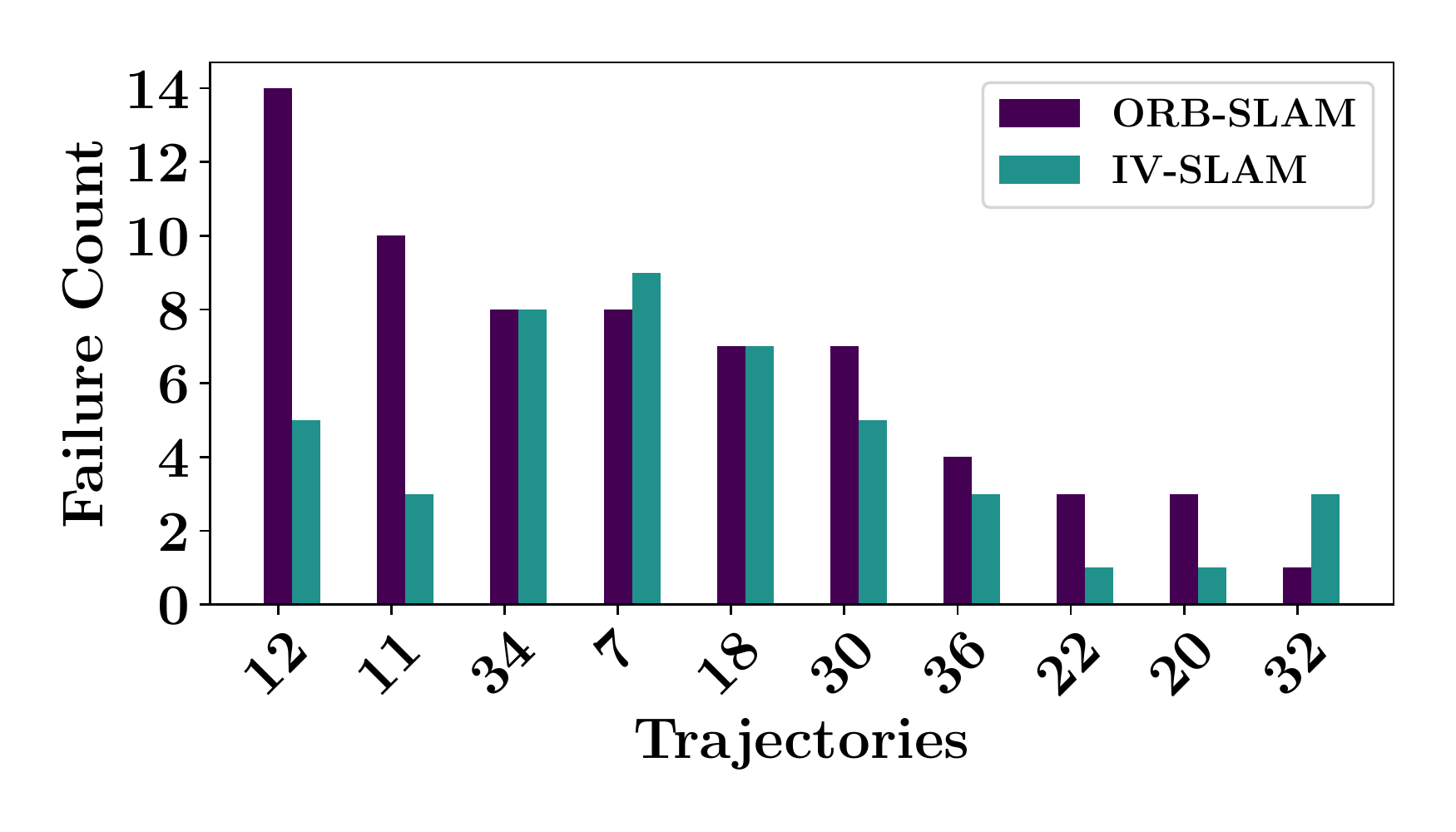}
    \caption{}
    \label{fig:failure_count_airsim}
  \end{subfigure}
  \begin{subfigure}[b]{0.32\linewidth}
    \includegraphics[width=1\linewidth, trim=0 0 0 0,clip]{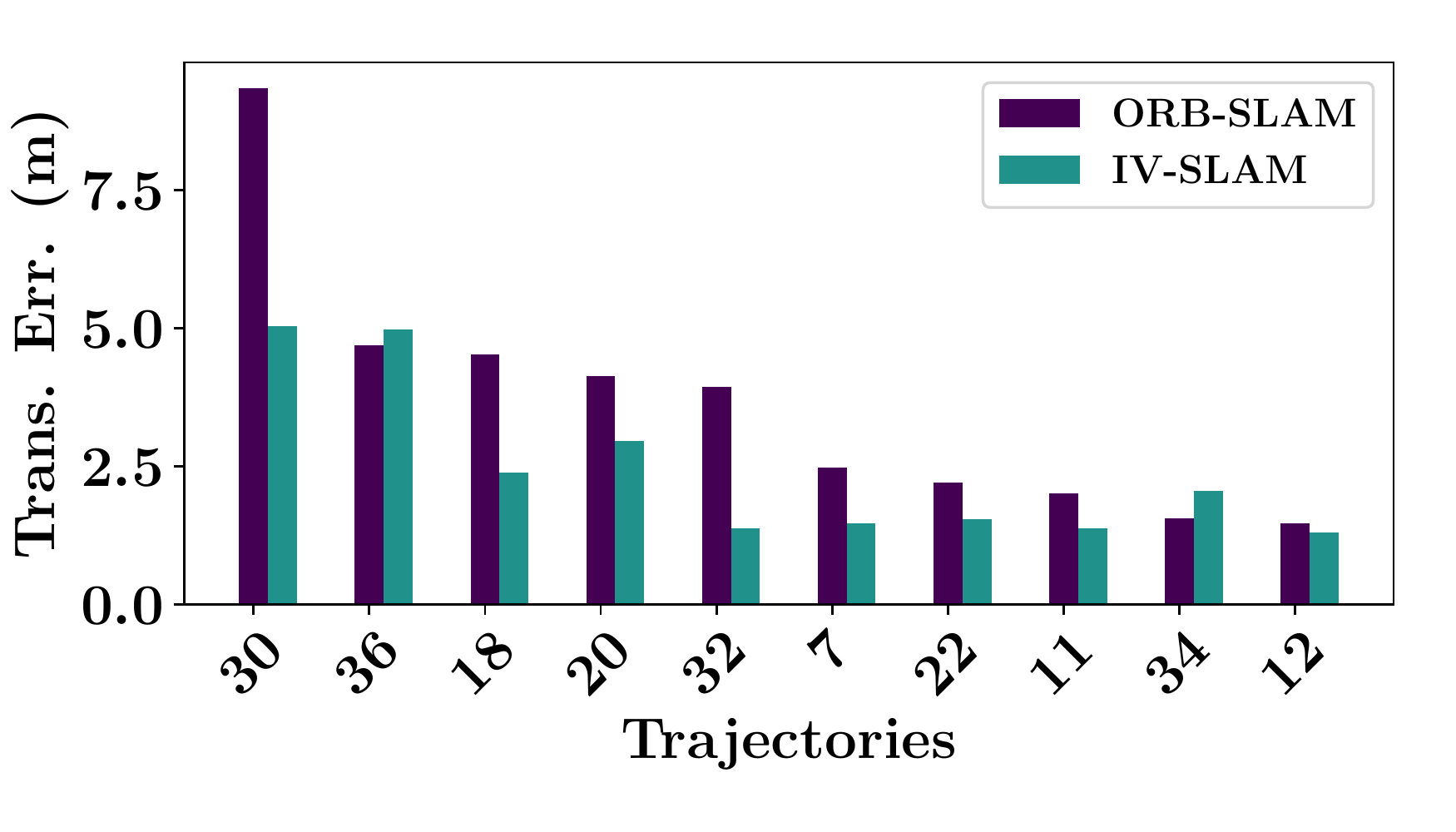}
    \caption{}
    \label{fig:rpe_trans_airsim}
  \end{subfigure}
  \begin{subfigure}[b]{0.32\linewidth}
    \includegraphics[width=1\linewidth, trim=0 0 0 0,clip]{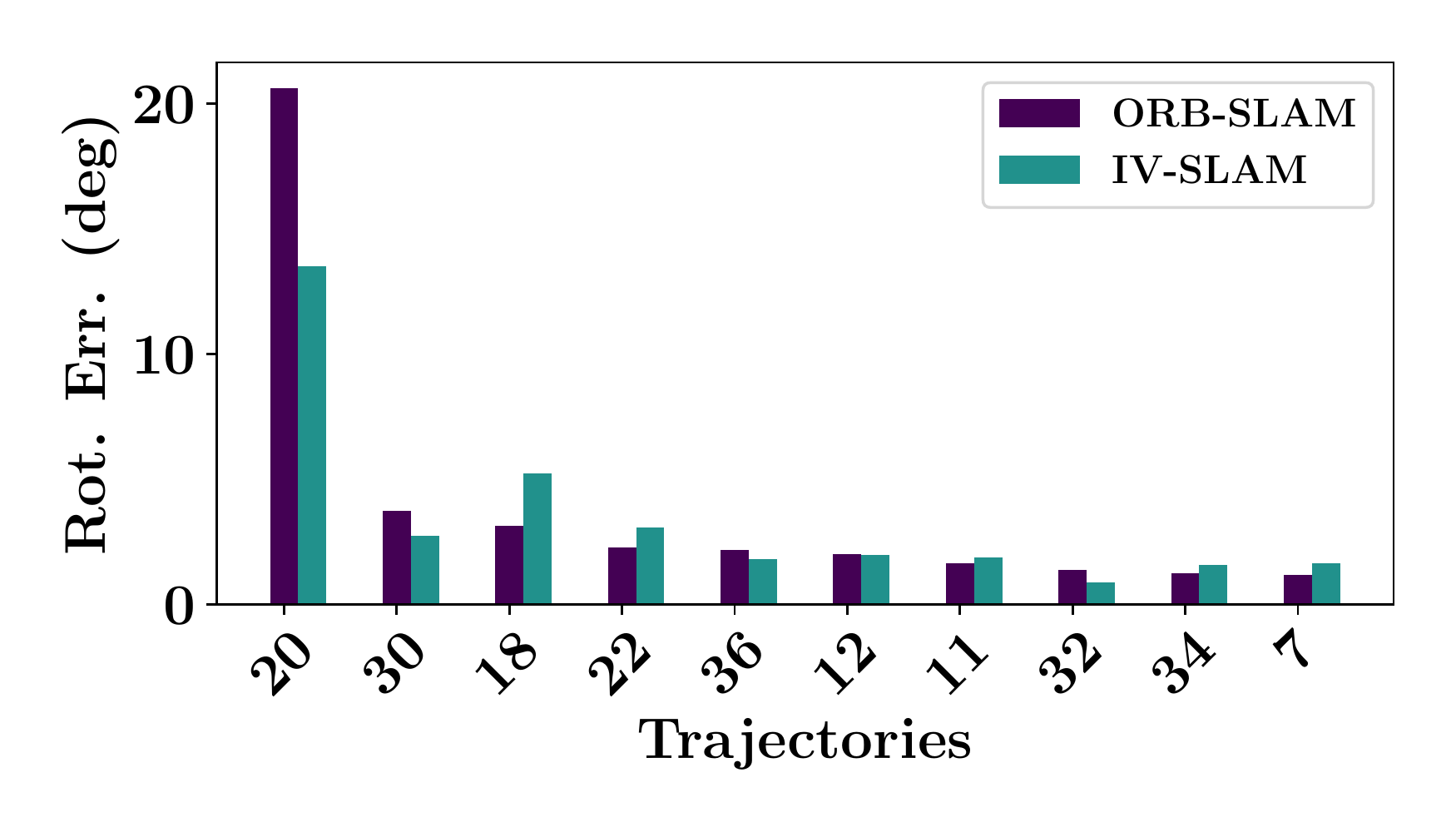}
    \caption{}
    \label{fig:rpe_rot_airsim}
  \end{subfigure}
  \caption{Per trajectory comparison of the performance of IV-SLAM and ORB-SLAM in the simulation experiment. (\subref{fig:failure_count_airsim}) Tracking failure count. (\subref{fig:rpe_trans_airsim}) RMSE of translational error and (\subref{fig:rpe_rot_airsim}) RMSE of rotational error over consecutive $20$\si{m}-long horizons.}
  \label{fig:per_traj_results_airsim}
  \vspace{-3mm}
 \end{figure}
 
 \begin{figure}[t]
  \begin{subfigure}[b]{0.32\linewidth}
    \includegraphics[width=1\linewidth, trim=0 0 0 0,clip]{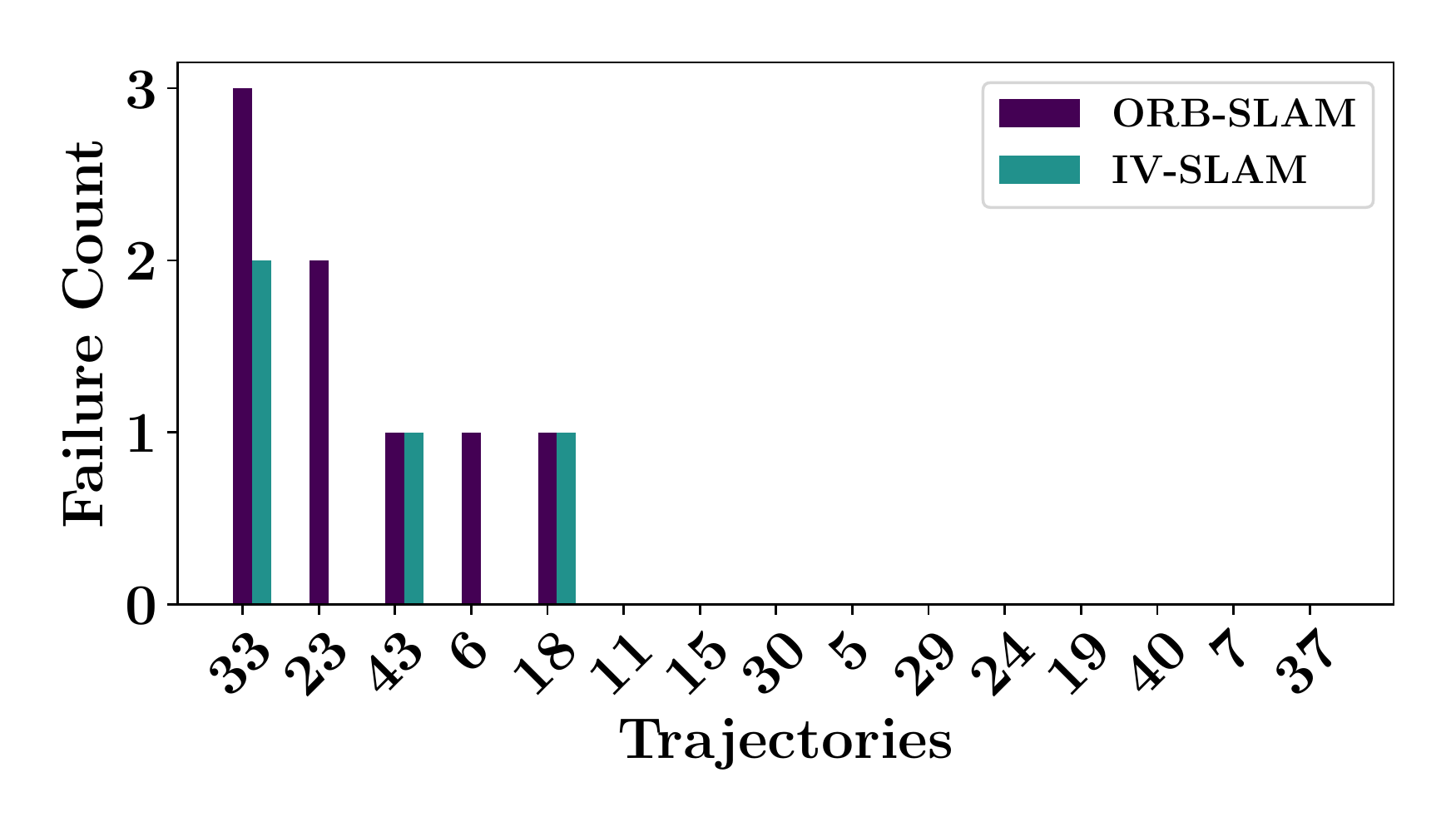}
    \caption{}
    \label{fig:failure_count_jackal}
  \end{subfigure}
  \begin{subfigure}[b]{0.32\linewidth}
    \includegraphics[width=1\linewidth, trim=0 0 0 0,clip]{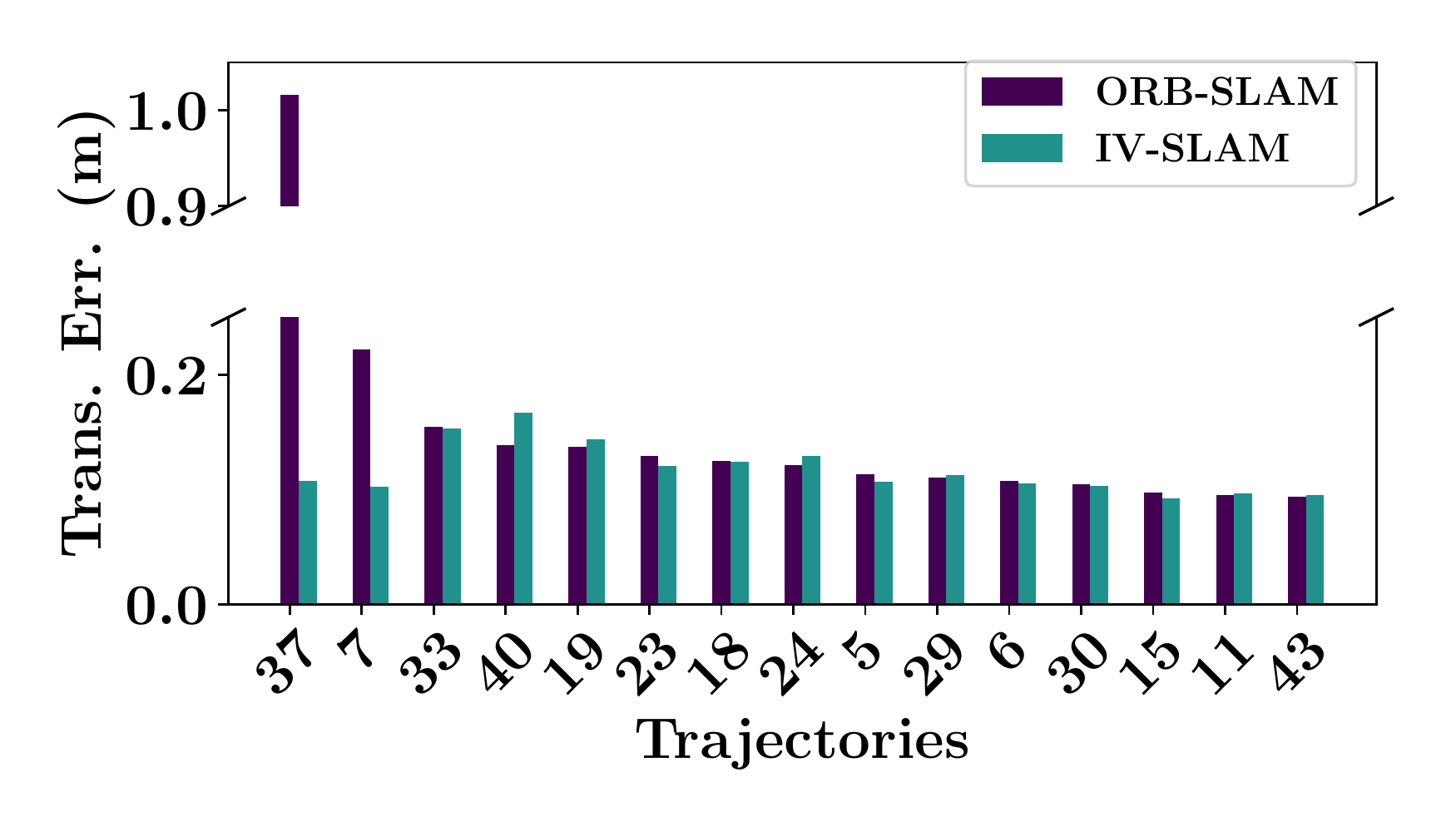}
    \caption{}
    \label{fig:rpe_trans_jackal}
  \end{subfigure}
  \begin{subfigure}[b]{0.32\linewidth}
    \includegraphics[width=1\linewidth, trim=0 0 0 0,clip]{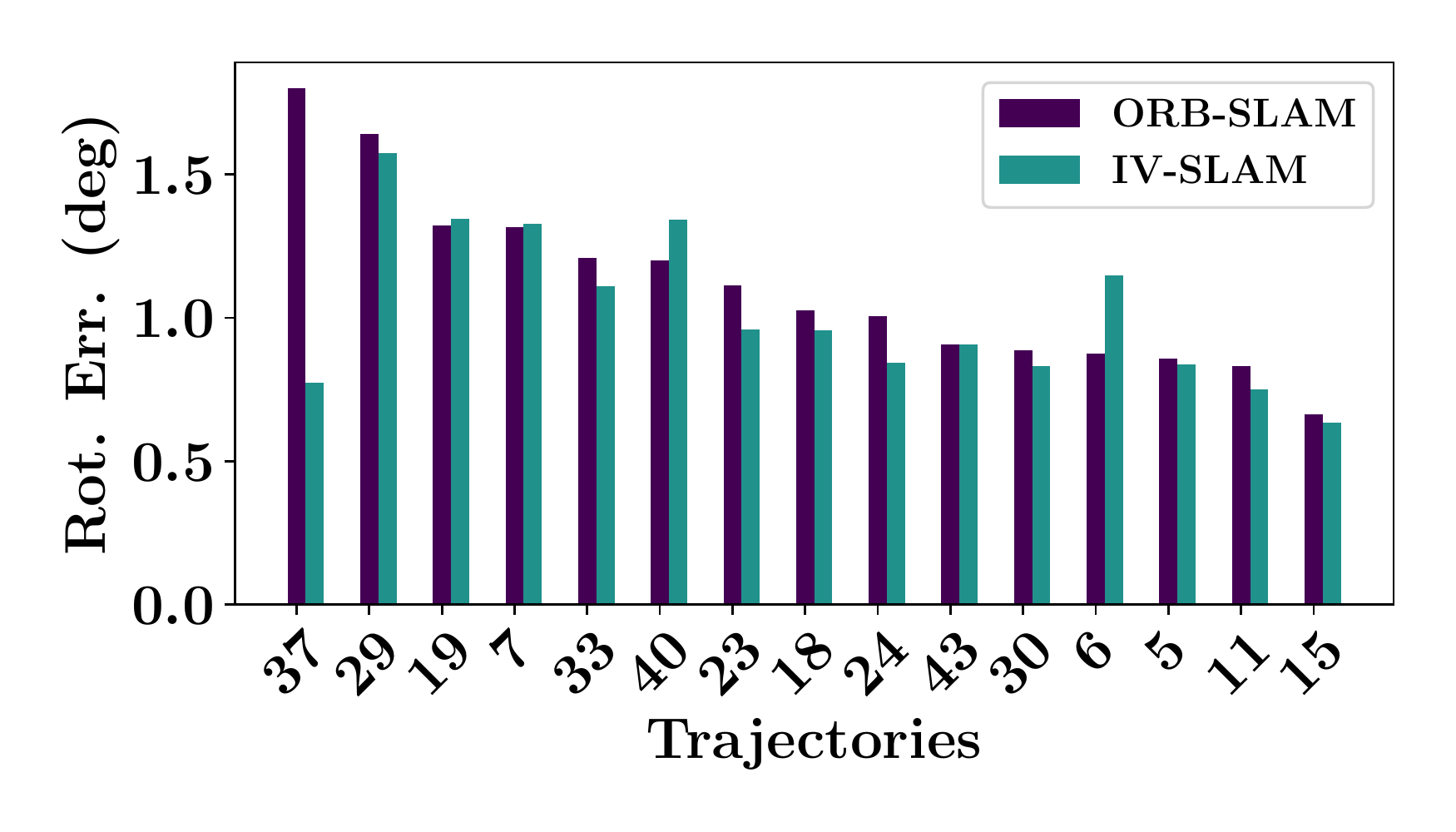}
    \caption{}
    \label{fig:rpe_rot_jackal}
  \end{subfigure}
  \caption{Per trajectory comparison of the performance of IV-SLAM and ORB-SLAM in the real-world experiment. (\subref{fig:failure_count_jackal}) Tracking failure count. (\subref{fig:rpe_trans_jackal}) RMSE of translational error and (\subref{fig:rpe_rot_jackal}) RMSE of rotational error over consecutive $2$\si{m}-long horizons.}
  \label{fig:per_traj_results_jackal}
  \vspace{-3mm}
 \end{figure}

 \begin{figure}[h]
   \centering
   \includegraphics[width=0.5\linewidth]{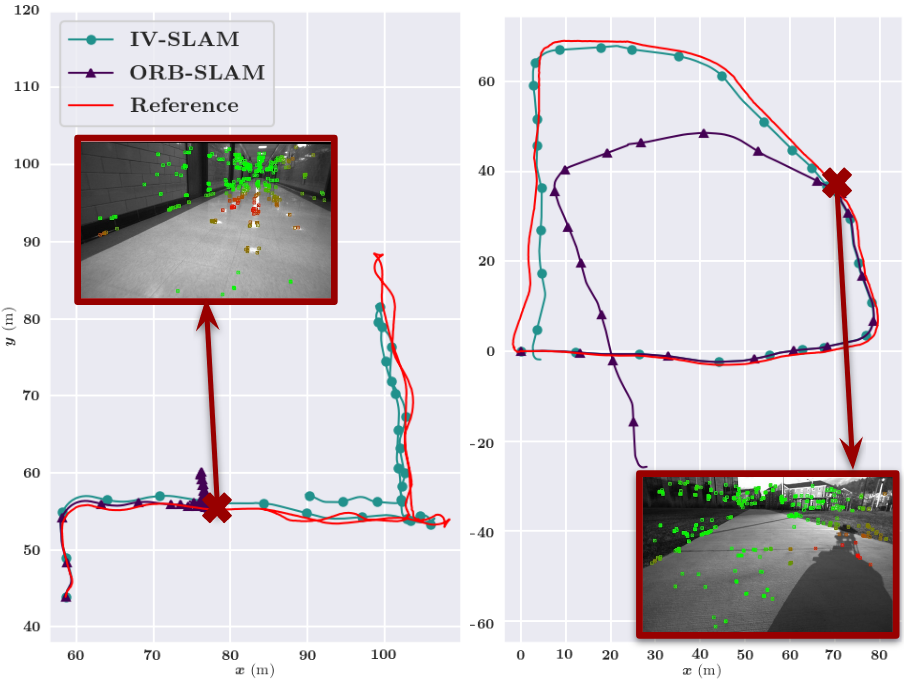}
   \caption{Example deployment sessions of the robot. IV-SLAM successfully follows the reference camera trajectory while ORB-SLAM leads to severe tracking errors caused by image features extracted on the shadow of the robot and surface reflections.}
    \label{fig:trajectory_example_jackal}
 \end{figure}

\paragraph{Challenging datasets}
We also evaluate IV-SLAM on the simulation and real-world datasets introduced in \cref{ivslam_dataset}, which include scenes that are representative of the challenging scenarios that can happen in the real-world applications of V-SLAM.
Given the larger scale of the environment, and faster speed of the robot in the simulation dataset, we pick $d_R=2$\si{m} for the real-world environment and $d_S=20$\si{m} for the simulation. Figures~\ref{fig:per_traj_results_airsim} and~\ref{fig:per_traj_results_jackal} compare the per trajectory tracking error values as well as the tracking failure count for IV-SLAM and ORB-SLAM in both experimental environments.
For most of the trajectories, IV-SLAM has a lower number of tracking failures. IV-SLAM also achieves similar or less RPE in most of the trajectories. 
RPE provides a metric of the local accuracy of the estimated pose of the robot and is calculated for each method only for parts of the trajectory where they have been able to track the pose of the robot. 
Since the baseline ORB-SLAM had more tracking failures, more parts of the trajectories, including the challenging sections where the failures happen, have been excluded when computing RPE for ORB-SLAM.
As a result, the lower number of tracking failures for IV-SLAM while keeping a similar or lower local accuracy level shows that it achieves better global accuracy across the different trajectories.\changes{In the simulation experiments, trajectories 7 and 32 exhibit more tracking failures in IV-SLAM. This is due to challenging high-speed turning maneuvers in these datasets, where there are very few matched features across consecutive frames. IV-SLAM rejects unreliable features such as those present on high-frequency asphalt textures near the robot, but it faces a limitation in replacing these rejected features with new matched features due to the significant change in the camera viewpoint during high-speed turns. On the other hand, the baseline algorithm does not remove any of the unreliable features, resulting in continuous tracking but suffering from high RPE, as illustrated in figure~\ref{fig:per_traj_results_airsim}.}
Table~\ref{table:aggregate_results} summarizes the results and shows the RMSE values calculated over all trajectories.
The results show that IV-SLAM leads to more than $70\%$ increase in the mean distance between failures (MDBF) and a $35\%$ decrease in the translation error in the real-world dataset. IV-SLAM similarly outperforms the original ORB-SLAM in the simulation dataset by both reducing the tracking error and increasing MDBF.
As it can be seen numerous tracking failures happen in both environments and the overall error rates are larger than those corresponding to the KITTI and EuRoC datasets due to the more difficult nature of these datasets. It is noteworthy that the benefit gained from using IV-SLAM is also more pronounced on these datasets with challenging visual settings.
Figure~\ref{fig:trajectory_example_jackal} demonstrates example deployment sessions of the robot from the real-world dataset and compares the reference pose of the camera with the estimated trajectories by both algorithms under test. It shows how image features extracted from the shadow of the robot and 
surface reflections cause significant tracking errors for ORB-SLAM, while IV-SLAM successfully handles such challenging situations.

\subsection{Categorizing Sources of Perception Errors} \label{sec:perception_error_clustering}
\begin{figure*}[t]
  \centering 
  \includegraphics[width=1\linewidth,trim=0 0 0 0,clip]{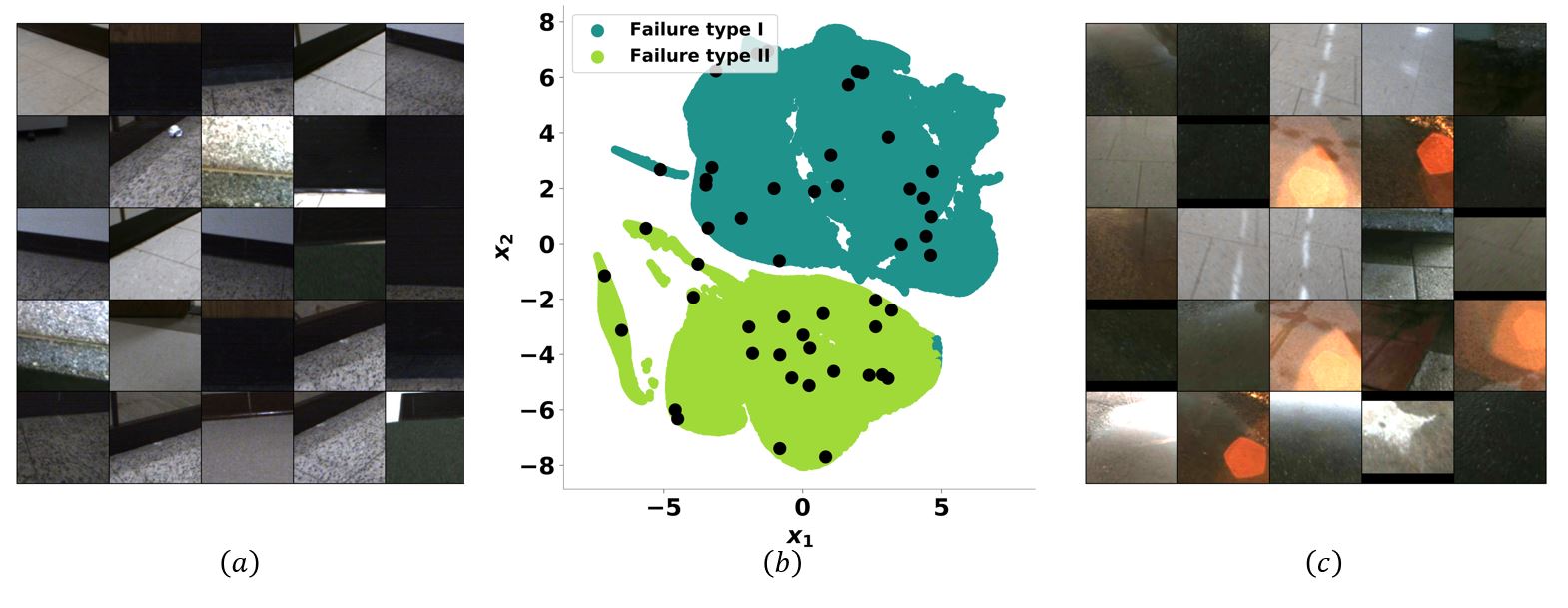}
  \caption{Clustering predicted sources of error using data embeddings learned by the introspection function of the stereo depth estimator. b) Visualization of the two extracted clusters of perception errors in the embedding space after being projected down to $2D$ via dimensionality reduction. a) Randomly sampled image patches from the failure type II cluster, shown as black dots on (b). c) Image patches randomly sampled from the failure type I cluster. Reflection and lens glare (type I) and the dark stripes at the bottom of the walls (type II) are the most dominant sources of error for the perception algorithm under test.}
  \label{fig:failure_case_clustering_wide}
\end{figure*}

Using deep learning for approximating the introspection function has the advantage of learning representations of the sensory data, which capture properties that pose challenges for the perception algorithms and are correlated with occurrence of different types of errors in perception. 
Such learned representations are useful for debugging and analysis of a perception algorithm and identifying causes of perception errors from logged data.
To test this hypothesis, we inspect the learned representation by the introspection function of the depth estimator. 
We run the introspection function on the test dataset and for all data-points predicted to lead to perception errors, we cluster the extracted embedding vectors using the k-means approach.
Figure~\ref{fig:failure_case_clustering_wide} illustrates the resultant clusters projected down to 2D using t-SNE~\citep{maaten2008visualizing} as well as sampled image patches from each cluster. The result shows that the dark edges at the bottom of the walls and reflection/glare to be the most dominant sources of error for the perception algorithm under test.

\subsection{Introspection for Fully Learned Perception}\label{sec:result_ipr_fully_learned_perc}

In order to evaluate the effectiveness of introspective perception for fully learned perception algorithms, we implement introspective perception for GA-Net~\cite{zhang2019ga}, a deep neural network for stereo depth estimation.
We compare our approach with SOTA uncertainty estimation methods that exist for deep learning models in terms of 1) failure prediction accuracy for in-distribution data, 2) failure prediction accuracy for out-of-distribution data, and 3) their computational load.
We use the simulation dataset described in \cref{sec:depth_dataset_sim} for our experiments.

We compare \IPrName{} against MCDropout and Ensemble, two common approaches for uncertainty estimation in deep learning. We evaluate them in terms of their ability to predict instances of depth estimation failures. 
We modify the GA-Net architecture to add dropout layers after ReLU activation units, and then train several instances of the network with different random initializations. For MCDropout, a single instance of the network is used and dropout is turned on during inference. We use a Monte Carlo sample size of 3. For Ensemble, multiple instances of the network are used and dropout is turned off during inference. We use 3 networks in our ensemble.
As explained in~\cref{{sec:approximate_BNN}} MCDropout and Ensemble include the variance of the output of each model in the ensemble (or each Monte Carlo samples) as a hyperparameter. We tune this hyperparameter to optimize for the negative log likelihood (NLL) on the training data. We also include in the experiments a version of both MCDropout and Ensemble without the hyperparameter tuning, which we refer to as MCDropout-Uncalib and Ensemble-Uncalib, respectively. 
The reason for including both calibrated and uncalibrated versions is to investigate whether the approximate BNN approach can be used without any training or calibration after the perception algorithm is trained and hence has the benefit of ease of use.

We train the GA-Net instances and the introspection function and also calibrate MCDropout and Ensemble on the same subset of the data from the City environment. The remaining data collected in the City environment, which is from novel trajectories of the robot navigating the environment under the same weather conditions present in the training dataset, comprises the in-distribution (ID) test dataset. Moreover, the data from the Neighborhood environment comprises the out-of-distribution (OOD) test dataset.

\subsubsection{Failure Prediction Accuracy for In-distribution Data}

\begin{table}[]
  \centering
  \resizebox{0.7\textwidth}{!}{
  \begin{tabular}{ccccl}
  \hline
  Method            & Precision       & Recall          & F-1 Score       & NLL               \\ \hline
  IPr               & 0.6333          & \textbf{0.8704} & 0.6874          & \textbf{-0.8717} \\
  Ensemble          & 0.6642          & 0.8525          & \textbf{0.7197} & -0.8182          \\
  Ensemble-Uncalib  & 0.8553          & 0.5721          & 0.6156          & -0.6191          \\
  MCDropout         & 0.5897          & 0.7998          & 0.6259          & -0.7853          \\
  MCDropout-Uncalib & \textbf{0.9875} & 0.5000          & 0.4937          & -0.6192          \\ \hline
  \end{tabular}
  }
  \caption{Failure Prediction Results for In-Distribution Data}
  \label{tab:failure_prediction_id_dataset}
\end{table}

We run the test data through GA-Net equipped with the introspection function, the GA-Net ensemble, as well as the GA-Net MCDropout. For a pair of input stereo images, each method outputs an estimated depth image along with an estimated uncertainty image. 
For each pixel in a predicted depth image, we label the prediction as a failure if the depth error is greater than a threshold of \SI{1.0}{\meter}. 
We quantify the accuracy of each uncertainty estimation method by comparing the ground truth failure labels with the per-pixel probability of failure predicted by each method. 
As explained in \cref{sec:approximate_BNN}, MCDropout and Ensemble estimate the standard deviation of the predicted depth values, approximating the predictions to be from Gaussian distributions. For each pixel in the predicted depth image, the probability of failure is calculated in terms of the estimated uncertainty $\sigma_{\text{ens}}$ as $p(\text{failure}) = 1 - \erf( \frac{1}{\sqrt{2} \sigma_{\text{ens}}})$.

Table~\ref{tab:failure_prediction_id_dataset} demonstrates the failure prediction results in terms of precision, recall, F-1 score, and negative log likelihood (NLL) for all the methods under test. 
\IPrName{} achieves the best results in terms of recall rate and NLL, and Ensemble achieves the highest F-1 score. The uncalibrated versions of both Ensemble and MCDropout perform poorly and exhibit overconfidence in the accuracy of the depth estimates. The uncalibrated MCDropout predicts almost no predictions to be failures. 

It should be noted that while precision, recall, and F1-score evaluate the classification performance of the different methods in predicting failures, the NLL metric also captures how well the predicted probabilities are distributed. In other terms, NLL evaluates how well the predicted error distribution matches the empirical error values. \IPrName{} achieves better NLL results since it is specifically trained to predict the error distribution of the perception algorithm under test, whereas the other methods report the empirical variance of the networks predictions as an approximate measure of uncertainty.

\subsubsection{Failure Prediction Accuracy for Out-of-distribution Data}
\begin{table}[]
  \centering
  \resizebox{0.7\textwidth}{!}{
  \begin{tabular}{ccccl}
  \hline
  Method            & Precision         & Recall           & F-1 Score        & NLL               \\ \hline
  IPr               & 0.6162            & 0.8157           & 0.6521          & -0.8056          \\
  IPr- Ensemble     & 0.6091           & \textbf{0.8310} & 0.6404          & \textbf{-0.8083} \\
  Ensemble          & 0.6730           & 0.8271          & \textbf{0.7203} & -0.7696          \\
  Ensemble-Uncalib  & 0.9704          & 0.5000           & 0.4879          & -0.5286           \\
  MCDropout         & 0.6312           & 0.8121         & 0.6724          & -0.7496          \\
  MCDropout-Uncalib & \textbf{0.9763} & 0.5000          & 0.4879         & -0.5285          \\ \hline
  \end{tabular}
  }
  \caption{Failure Prediction Results for Out-Of-Distribution Data}
  \label{tab:failure_prediction_ood_dataset}
  \end{table}

We separately evaluate the performance of the different methods in predicting failures of perception for data that is out-of-distribution with respect to the environments used for training both the perception algorithm (GA-Net) and the introspection function. 
In addition to the methods that we tested for the ID dataset, here we also implement and test a combination of \IPrName{} and Ensemble. We train an ensemble of 3 introspection functions for GA-Net and use the average output of the functions for predicting probability of failures. Our motivation for this is to investigate whether the ensemble approach can be used to estimate the uncertainty of the introspection function itself for OOD data.

Table~\ref{tab:failure_prediction_ood_dataset} shows the failure prediction results summary for the OOD dataset. 
As expected, the failure prediction accuracy is reduced and NLL increased for all methods compared to the ID dataset. However, similar to the ID dataset, the uncalibrated versions of Ensemble and MCDropout are overconfident, Ensemble outperforms MCDropout, and \IPrName{} outperforms both in terms of NLL. 
The IPr-Ensemble achieves a marginally better NLL compared to \IPrName{}. 
The results show that \IPrName{} does a good job of predicting errors of perception also on OOD data, and that leveraging the idea of ensembles in the implementation of the introspection function leads to better or similar performance compared to \IPrName{} alone in novel environments.

\subsubsection{Computational Load and Memory Consumption}

Robots are limited by the computational resources available to them, hence it is important to evaluate the additional computational cost incurred by performing uncertainty estimation. We measure the GPU inference time and memory consumption for each of the uncertainty estimation methods during inference. Table~\ref{tab:gpu_inference_time_memory} summarizes the results.
\IPrName{} is more than $15$ times faster while consuming significantly less GPU memory compared to MCDropout and Ensemble. Since \IPrName{} tackles the problem of predicting the error distribution of the perception algorithm as opposed to solving the full perception problem, it can be implemented with a significantly smaller network architecture, making it ideal for real-time robotic applications, where the computational resources are limited.

\begin{table}[]
  \centering
  \resizebox{0.5\textwidth}{!}{
  \begin{tabular}{@{}ccc@{}}
  \toprule
  Model     & \begin{tabular}[c]{@{}c@{}}Inference Time\\ (ms)\end{tabular} & Memory (MB)  \\ \midrule
  IPr       & \textbf{16.70 $\pm$ 0.11}                                         & \textbf{16.97 $\pm$ 0.00} \\
  Ensemble  & 244.85 $\pm$ 1.91                                                & 222.00 $\pm$ 0.00   \\
  MCDropout & 244.80 $\pm$ 1.92                                                & 222.00 $\pm$ 0.00   \\ \bottomrule
  \end{tabular}
  }
  \caption{GPU Inference Time and Memory Usage}
  \label{tab:gpu_inference_time_memory}
\end{table}

\subsubsection{Qualitative Resutls}

\begin{figure}[t]
  \begin{subfigure}[b]{0.161\linewidth}
    \caption*{\scriptsize Ground Truth}
    \includegraphics[width=1.05\linewidth, trim=0 0 0 0,clip]{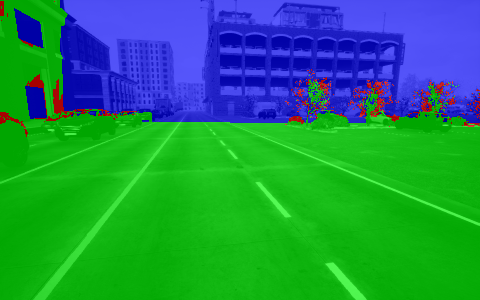}
  \end{subfigure} 
  \begin{subfigure}[b]{0.161\linewidth}
    \caption*{\centering \scriptsize MCDropout-Uncalibrated}
    \includegraphics[width=1.05\linewidth, trim=0 0 0 0,clip]{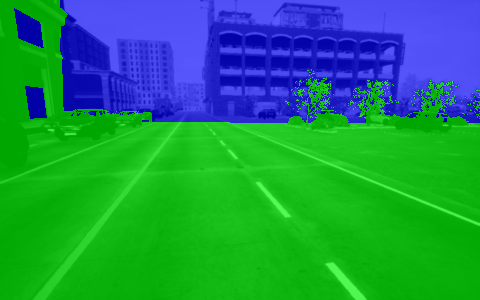}
  \end{subfigure} 
  \begin{subfigure}[b]{0.161\linewidth}
    \caption*{\centering \scriptsize Ensemble-Uncalibrated}
    \includegraphics[width=1.05\linewidth, trim=0 0 0 0,clip]{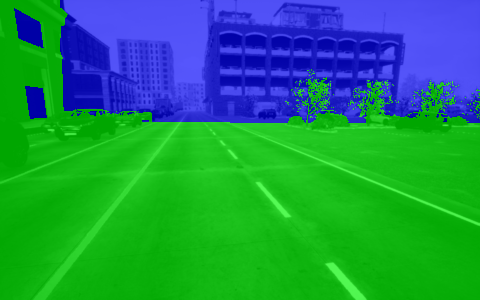}
  \end{subfigure} 
  \begin{subfigure}[b]{0.161\linewidth}
    \caption*{\scriptsize MCDropout}
    \includegraphics[width=1.05\linewidth, trim=0 0 0 0,clip]{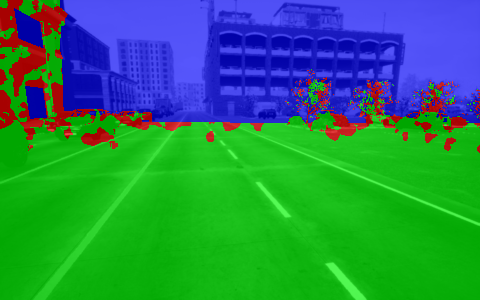}
  \end{subfigure}
  \begin{subfigure}[b]{0.161\linewidth}
    \caption*{\scriptsize Ensemble}
    \includegraphics[width=1.05\linewidth, trim=0 0 0 0,clip]{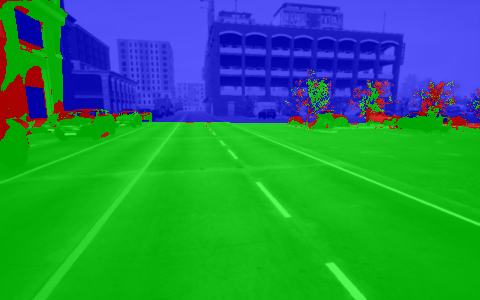}
  \end{subfigure}
  \begin{subfigure}[b]{0.161\linewidth}
    \caption*{\scriptsize \IPrName{}}
    \includegraphics[width=1.05\linewidth, trim=0 0 0 0,clip]{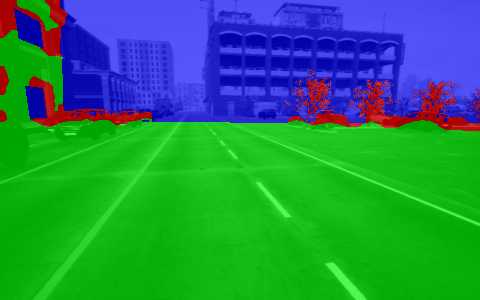}
  \end{subfigure}

  \begin{subfigure}[b]{0.161\linewidth}
    \includegraphics[width=1.05\linewidth, trim=0 0 0 0,clip]{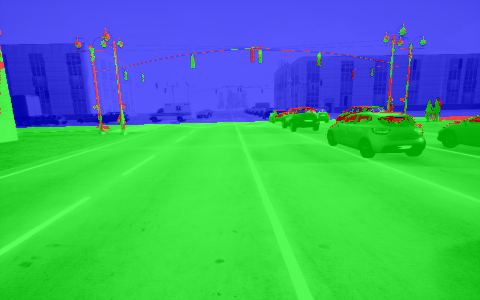}
  \end{subfigure} 
  \begin{subfigure}[b]{0.161\linewidth}
    \includegraphics[width=1.05\linewidth, trim=0 0 0 0,clip]{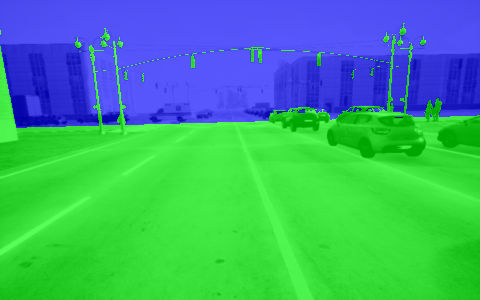}
  \end{subfigure} 
  \begin{subfigure}[b]{0.161\linewidth}
    \includegraphics[width=1.05\linewidth, trim=0 0 0 0,clip]{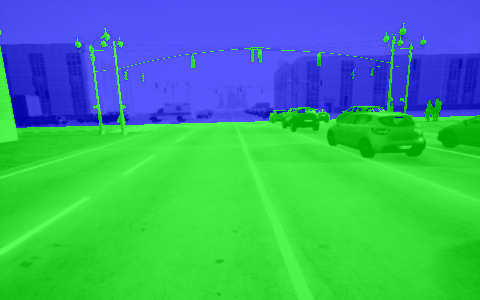}
  \end{subfigure} 
  \begin{subfigure}[b]{0.161\linewidth}
    \includegraphics[width=1.05\linewidth, trim=0 0 0 0,clip]{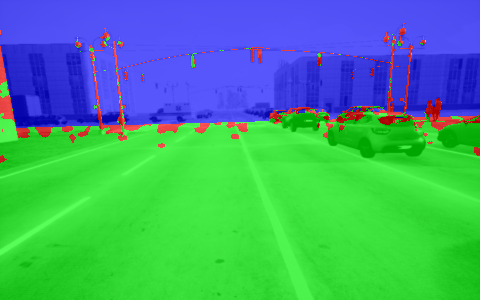}
  \end{subfigure}
  \begin{subfigure}[b]{0.161\linewidth}
    \includegraphics[width=1.05\linewidth, trim=0 0 0 0,clip]{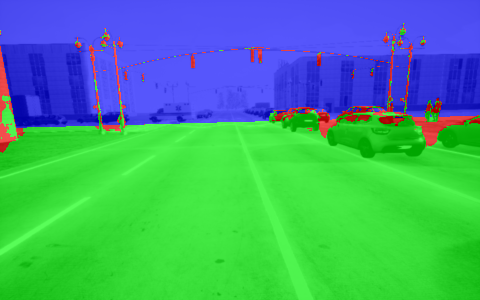}
  \end{subfigure}
  \begin{subfigure}[b]{0.161\linewidth}
    \includegraphics[width=1.05\linewidth, trim=0 0 0 0,clip]{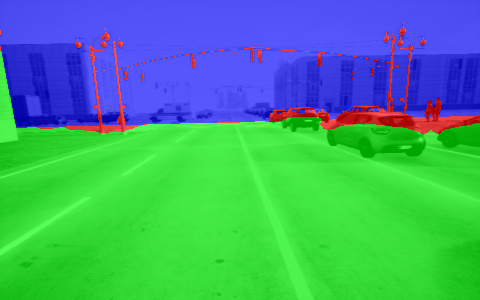}
  \end{subfigure}

  \begin{subfigure}[b]{0.161\linewidth}
    \includegraphics[width=1.05\linewidth, trim=0 0 0 0,clip]{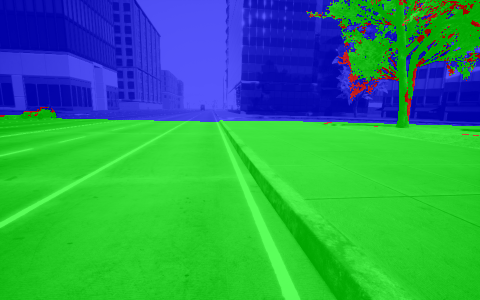}
  \end{subfigure} 
  \begin{subfigure}[b]{0.161\linewidth}
    \includegraphics[width=1.05\linewidth, trim=0 0 0 0,clip]{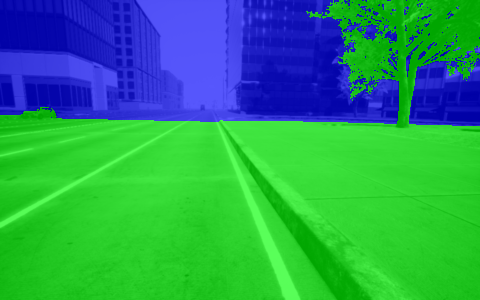}
  \end{subfigure} 
  \begin{subfigure}[b]{0.161\linewidth}
    \includegraphics[width=1.05\linewidth, trim=0 0 0 0,clip]{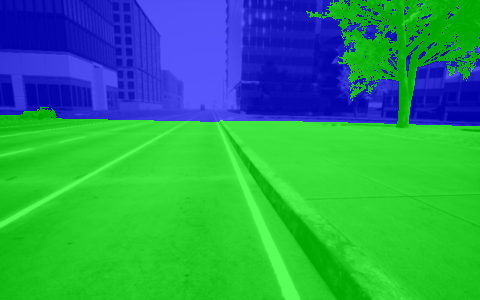}
  \end{subfigure} 
  \begin{subfigure}[b]{0.161\linewidth}
    \includegraphics[width=1.05\linewidth, trim=0 0 0 0,clip]{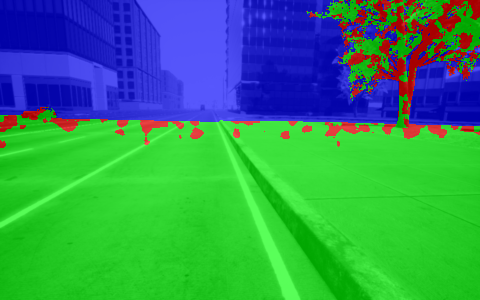}
  \end{subfigure}
  \begin{subfigure}[b]{0.161\linewidth}
    \includegraphics[width=1.05\linewidth, trim=0 0 0 0,clip]{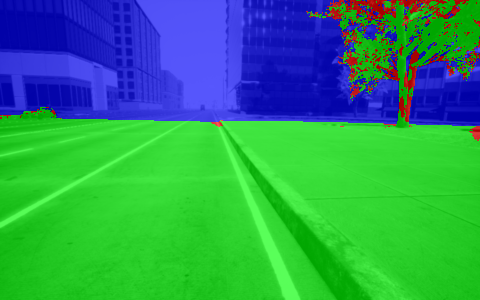}
  \end{subfigure}
  \begin{subfigure}[b]{0.161\linewidth}
    \includegraphics[width=1.05\linewidth, trim=0 0 0 0,clip]{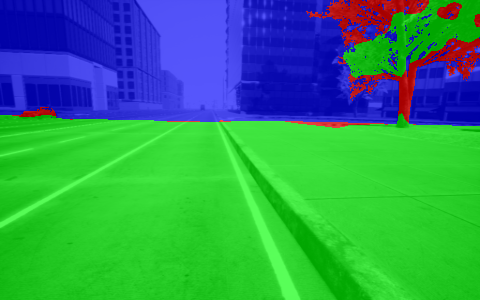}
  \end{subfigure}
 
  \caption{Visualizations of the failures of an end-to-end depth estimation network (GA-Net) and the predicted failures by different uncertainty prediction methods. 
  The input image to the depth estimator is colorized to highlight pixels, where the estimated depth is predicted to be correct (green), and pixels where the depth error is predicted to be larger than a threshold (red). Pixels that are colored blue are out of range.
  Both \IPrName{} and Ensemble correctly predict foliage, traffic lights, and edges of cars to lead to depth estimate failures.
}
  \label{fig:GA-Net_failure_snapshots}
 \end{figure}

In order to better understand the types of failures of the depth estimator under test, and how \IPrName{} and the other methods perform in terms of predicting these failures, we visualize sample qualitative results. Figure~\ref{fig:GA-Net_failure_snapshots} illustrates the ground truth depth estimate failures of GA-Net overlaid on the input images and compares that with predicted failures by the different methods. The uncalibrated versions of MCDropout and Ensemble predict almost no failures. MCDropout has erroneous predictions of failures in regions where the perception algorithm seems to consistently predict the correct depth values. \IPrName{} and Ensemble are both able to correctly pick up the failures of the depth estimator without significant false positives. They predict obstacles such as poles, trees, and edges of the cars to lead to depth estimate failures.

\section{Discussion and Future Work}\label{sec:discussion}

\paragraph*{Applying introspective perception to new perception tasks}
Introspective perception is intended to capture a wide class of perception algorithms regardless of the functional form. \IPrName{} applies to both machine learning and non-machine learning perception algorithms.
Given the vast range of perception algorithms, the introspective perception formulation does not provide a single algorithmic solution for all problems. Instead, it provides a general framework with a guideline for enabling a
mobile robot to learn an empirical approximation of the uncertainty of its
perception algorithms in the deployment environment and without the need
for human supervision.
Applying introspective perception to different perception tasks includes application-specific design choices such as the choice of the function approximator for the introspection function and the parametrization of the error distribution of perception.
We introduce two general classes of strategies to formulate self-supervision of introspective perception, and we believe that these two classes can provide insights into how to deploy \IPrName{} for new perception problems given their specific settings.

\paragraph*{The cost of collecting examples of perception errors}
\IPrName{} collects samples of errors of perception functions in the deployment environment in order to train a machine learning model that predicts the uncertainty of perception and hence prevents catastrophic failures of the robot.
While \IPrName{} requires a large amount of data to train an accurate and generalizable model, it does not require examples of a large number of catastrophic failures to train such a model.
Catastrophic failures are rare events in well-developed systems, however, small errors that do not always lead to catastrophic failures happen more frequently. Being able to learn to predict instances of smaller errors, allows introspective perception to predict and prevent catastrophic failures that are caused by the accumulation of small errors. For instance, during the training of IV-SLAM the introspection function is provided with frequent instances of large re-projection errors on the moving shadow of the robot, while these errors do not lead to catastrophic tracking failures in the existence of sufficient reliable data correspondences in the same scene. However, learning about the higher uncertainty in the re-projection error of image features extracted on the shadow of the robot and leveraging this information in the V-SLAM back-end prevents catastrophic tracking failures that happen when the majority of image features are extracted from the moving shadow of the robot.

\paragraph*{Correlation of errors of the perception function and the introspection function}
For cases when the perception function is an end-to-end machine learning algorithm, it is questionable whether the introspection function is susceptible to the same errors as the perception function since the input data to both the perception function and the introspection function is the same and both functions are implemented as machine learning models.
However, it should be noted that while the input data to the two functions is the same, the tasks of the two functions are different. 
The task of the introspection function is to estimate the aleatoric uncertainty of the perception function, i.e. predict the uncertainties in the output of the perception function that can not be reduced by providing more training data to the perception function. 
A machine learning perception algorithm might provide erroneous results when there is insufficient information in the input, e.g. when the input image is too dark for reliable depth estimation, however, the introspection function tackles the easier learning task of identifying that there is insufficient information in the input.

\paragraph*{Fusion of multiple introspection functions}
Some perception functions can be written as the composition of smaller perception functions.
For instance, as also shown in this work, V-SLAM is composed of a front-end that is responsible for feature detection and matching and a back-end that is responsible for estimating the pose of the camera given data associations provided by the front-end. In this work, we focused on developing an autonomously supervised method for learning an accurate estimate of the error distribution of each of the sub-components of a large perception algorithm that can each be defined as a perception function.
To estimate the error distribution of a complex perception function, we also need to reason about the propagation of errors in the output of the sub-components of the perception function.
For a complex perception algorithm,
modeling how errors in the output of one perception function, as estimated by its introspection function, propagate through a downstream perception function can be non-trivial, especially in the case of nonlinear perception functions. 
Extending the introspection function to leverage the uncertainty of the intermediate state variables as one of its inputs or using classical filtering methods~\citep{julier2004unscented} are potential approaches to tackle this problem.

\changes{
\paragraph*{The role of high-fidelity sensors in training of introspective perception}
While the availability of high-fidelity sensors is not a strict requirement for introspective perception, it is one of the viable methods to train the introspection function by leveraging consistency across sensors, as explained in~\cref{sec:ipr_training_method} and similar to the example of introspective stereo depth estimator presented in~\cref{sec:ivoa}. In situations where high-fidelity sensors are unavailable, we can use offline supervisory perception functions that consume the same sensory input as the primary perception function but provide higher accuracy outputs at the cost of increased computational requirements. For example, an offline depth estimator based on Neural Radiance Fields (NeRF)~\citep{kundu2022panoptic} can be employed to supervise a stereo depth estimator, although real-time deployment is not feasible.
Moreover, the use of spatio-temporal consistency constraints, as presented in~\cref{sec:ipr_training_method}, enables training of introspective perception without the necessity of a supervisory perception function or high-fidelity sensors.
}

\paragraph*{Future work}
There are several avenues for future work. First, \IPrName{} can be implemented for other safety-critical perception algorithms such as object detection, tracking, and lidar-based localization. 
Second, the introspection function can be generalized to leverage other informative data such as the robot control commands along with the perception inputs for predicting the errors of perception.
Third, in this work we employed a passive approach for collecting the training data for \IPrName{}, where all the data logged by the robot during deployment was used for training. However, \IPrName{} can be equipped with active-learning techniques~\citep{bajcsy1988active, krainin2011autonomous}
such that it guides the robot planner to collect more informative data online, and it subsamples a representative set of the logged data offline to train the introspection function more efficiently.
Moreover, in \cref{sec:perception_error_clustering}, we showed that representation of the sensor data learned by the introspection function can be used to cluster the perception errors that are due to the same source. It would be interesting to further investigate the performance of \IPrName{} in classifying the different sources of errors. Specifically, labeling and tracking the sources of errors in the test data in order to quantitatively evaluate IPr for identifying the different types of perception errors would be a great next step.

\section{Conclusion}\label{sec:conclusion}

In this paper, we defined and presented a general theory for introspective perception. We proposed autonomously supervised methods for training \IPrName{} in the deployment environment of a mobile robot. We provided examples of implementing \IPrName{} for two different perception algorithms: visual simultaneous localization and mapping and stereo depth estimation. We demonstrated that \IPrName{} reduces failures of autonomy that are associated with perception errors. 
Furthermore, we showed that \IPrName{} applies to both model-based and fully-learned perception algorithms, and is computationally significantly less expensive compared to the uncertainty estimation methods for deep learning based perception.

\section{Acknowledgments}
This work is supported in part by NSF (CAREER-2046955,
IIS-1954778) and DARPA (HR001120C0031). The views and conclusions contained in this document are
those of the authors only.

\bibliography{mybibfile}

\end{document}